\documentclass{article}

\usepackage{microtype}
\usepackage{graphicx}
\usepackage{booktabs}
\usepackage{times}
\usepackage{graphicx}
\usepackage{dsfont}
\usepackage{amsmath}
\usepackage{url}
\usepackage{amssymb}
\usepackage{color}
\usepackage{tabularx}
\usepackage{algorithm}
\usepackage[noend]{algorithmic}
\usepackage{capt-of,etoolbox}
\usepackage{algorithm}
\usepackage[noend]{algorithmic}
\usepackage{subfig} 
\usepackage{enumitem}
\usepackage[export]{adjustbox}
\usepackage{float}
\usepackage{arydshln}
\usepackage{hyperref}

\makeatletter
\patchcmd\@maketitle\null{{\myfigure{}\par}}{}{}
\g@addto@macro\normalsize{%
\addtolength{\abovedisplayskip}{-3pt}
\addtolength{\belowdisplayskip}{-3pt}
\addtolength{\abovedisplayshortskip}{-3pt}
\addtolength{\belowdisplayshortskip}{-3pt}
}
\usepackage[compact]{titlesec}
\titlespacing{\section}{0pt}{*0}{*0}
 \titlespacing{\subsection}{0pt}{*0}{*0}
\titlespacing{\subsubsection}{0pt}{*0}{*0}

\DeclareMathOperator*{\argmax}{arg\,max}

\usepackage[accepted]{icml2021}

\icmltitlerunning{Actionable Models: Unsupervised Offline Reinforcement Learning of Robotic Skills}

\begin{document}

\twocolumn[
\icmltitle{Actionable Models: \\Unsupervised Offline Reinforcement Learning of Robotic Skills}

\begin{icmlauthorlist}
\icmlauthor{Yevgen Chebotar}{google}
\icmlauthor{Karol Hausman}{google}
\icmlauthor{Yao Lu}{google}
\icmlauthor{Ted Xiao}{google}
\icmlauthor{Dmitry Kalashnikov}{google}
\icmlauthor{Jake Varley}{google}\\
\icmlauthor{Alex Irpan}{google}
\icmlauthor{Benjamin Eysenbach}{google,cmu}
\icmlauthor{Ryan Julian}{google,usc}
\icmlauthor{Chelsea Finn}{google,stanford}
\icmlauthor{Sergey Levine}{google,ucb}
\end{icmlauthorlist}

\icmlaffiliation{google}{Robotics at Google}
\icmlaffiliation{ucb}{UC Berkeley}
\icmlaffiliation{stanford}{Stanford University}
\icmlaffiliation{cmu}{Carnegie Mellon University}
\icmlaffiliation{usc}{University of Southern California}

\icmlcorrespondingauthor{Yevgen Chebotar}{chebotar@google.com}

\icmlkeywords{Robotics, Reinforcement Learning, Goal-Conditioned Learning, Unsupervised Learning, Self-Supervised Learning}

\vskip 0.3in
]

\printAffiliationsAndNotice{} 

\begin{abstract}
We consider the problem of learning useful robotic skills from previously collected offline data without access to manually specified rewards or additional online exploration, a setting that is becoming increasingly important for scaling robot learning by reusing past robotic data. In particular, we propose the objective of learning a functional understanding of the environment by learning to reach any goal state in a given dataset. We employ goal-conditioned Q-learning with hindsight relabeling and develop several techniques that enable training in a particularly challenging offline setting. We find that our method can operate on high-dimensional camera images and learn a variety of skills on real robots that generalize to previously unseen scenes and objects. We also show that our method can learn to reach long-horizon goals across multiple episodes through goal chaining, and learn rich representations that can help with downstream tasks through pre-training or auxiliary objectives. The videos of our experiments can be found at \url{https://actionable-models.github.io}
\end{abstract}


\begin{figure}[ht]
    \centering
    \vspace{-2pt}
    \includegraphics[trim=0.8cm 2.1cm 13.2cm 1.35cm, clip=true, width=0.95\columnwidth]{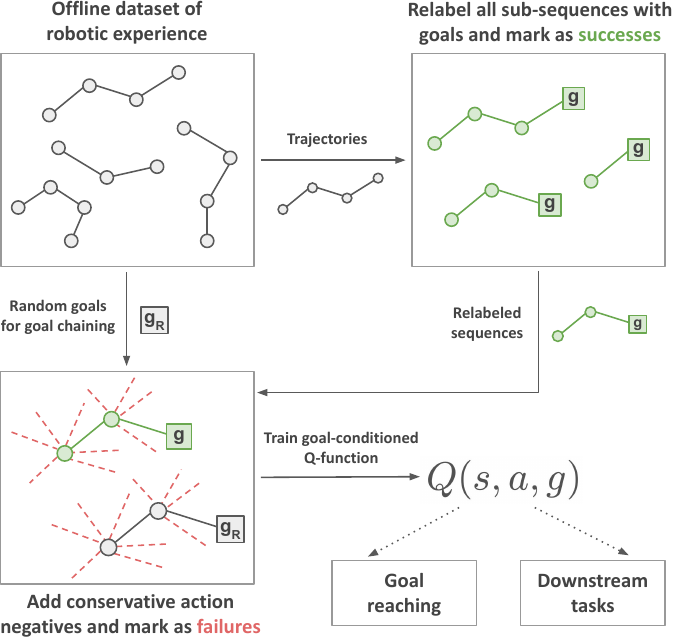}\\
    \vspace{4pt}
     \includegraphics[trim=11cm 0cm 31cm 2cm, clip=true, width=0.17\columnwidth]{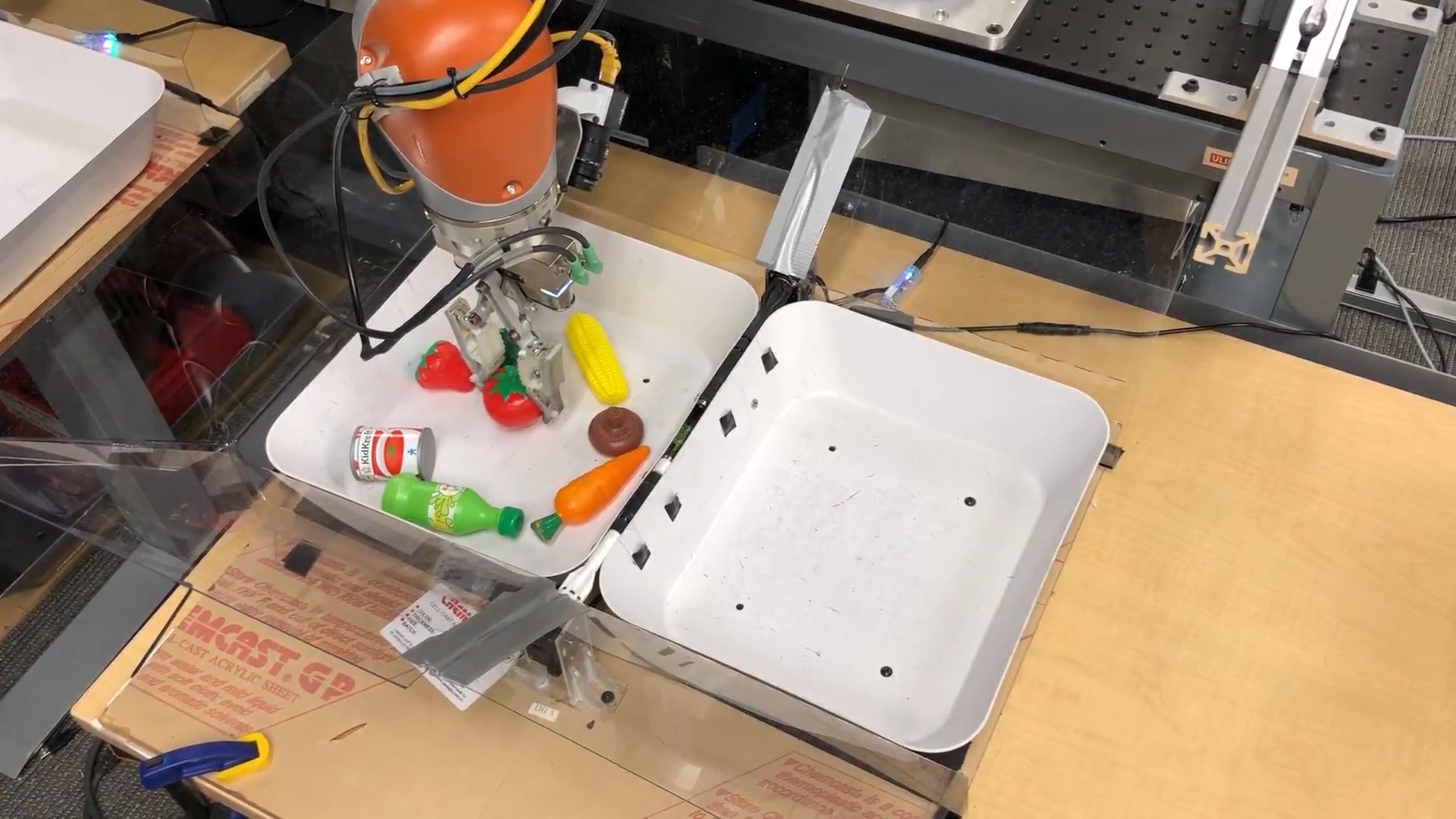} 
    \includegraphics[trim=15cm 1cm 17cm 6cm, clip=true, width=0.274\columnwidth]{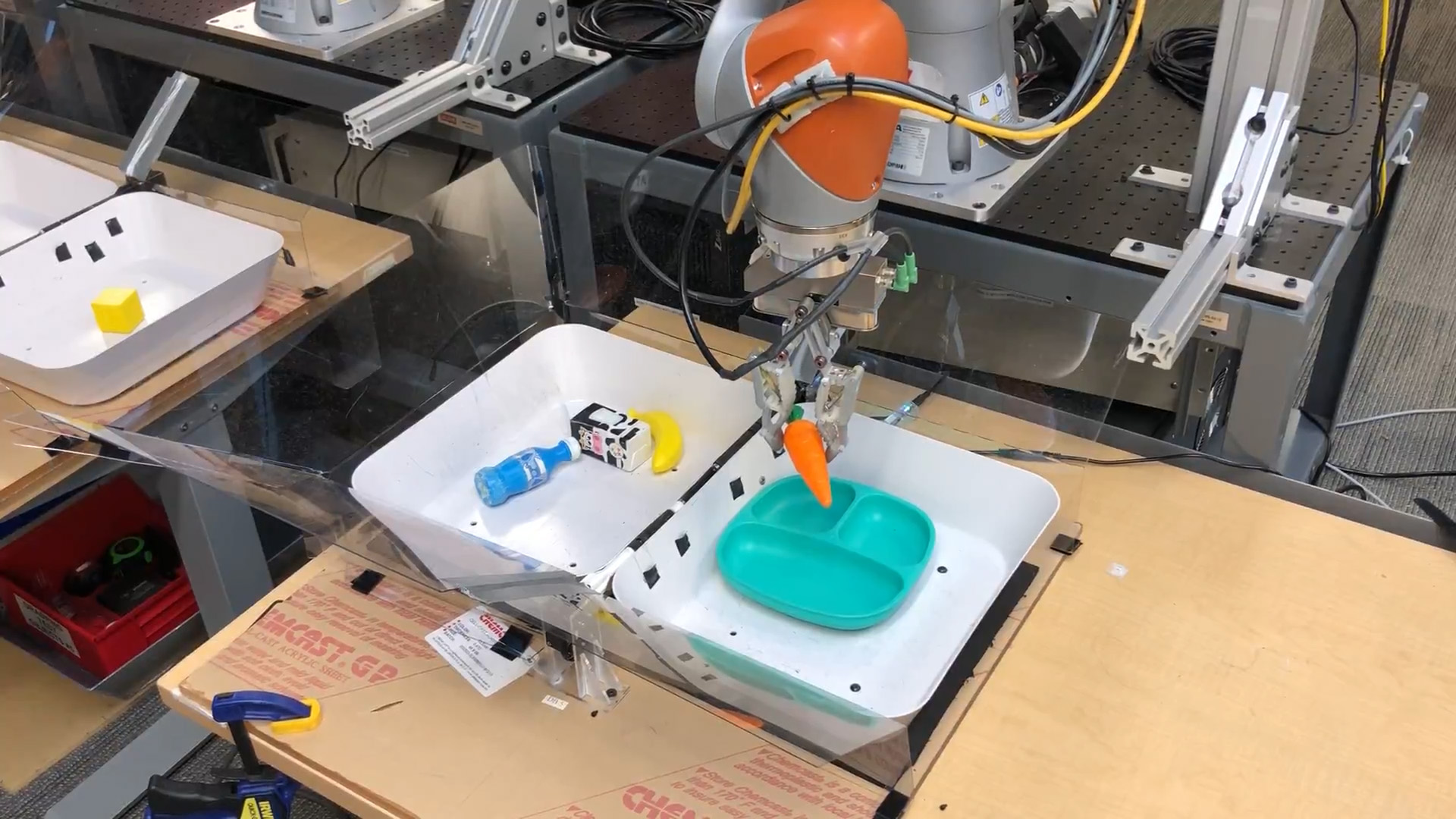}
    \includegraphics[trim=11cm 1.2cm 30cm 0cm, clip=true, width=0.173\columnwidth]{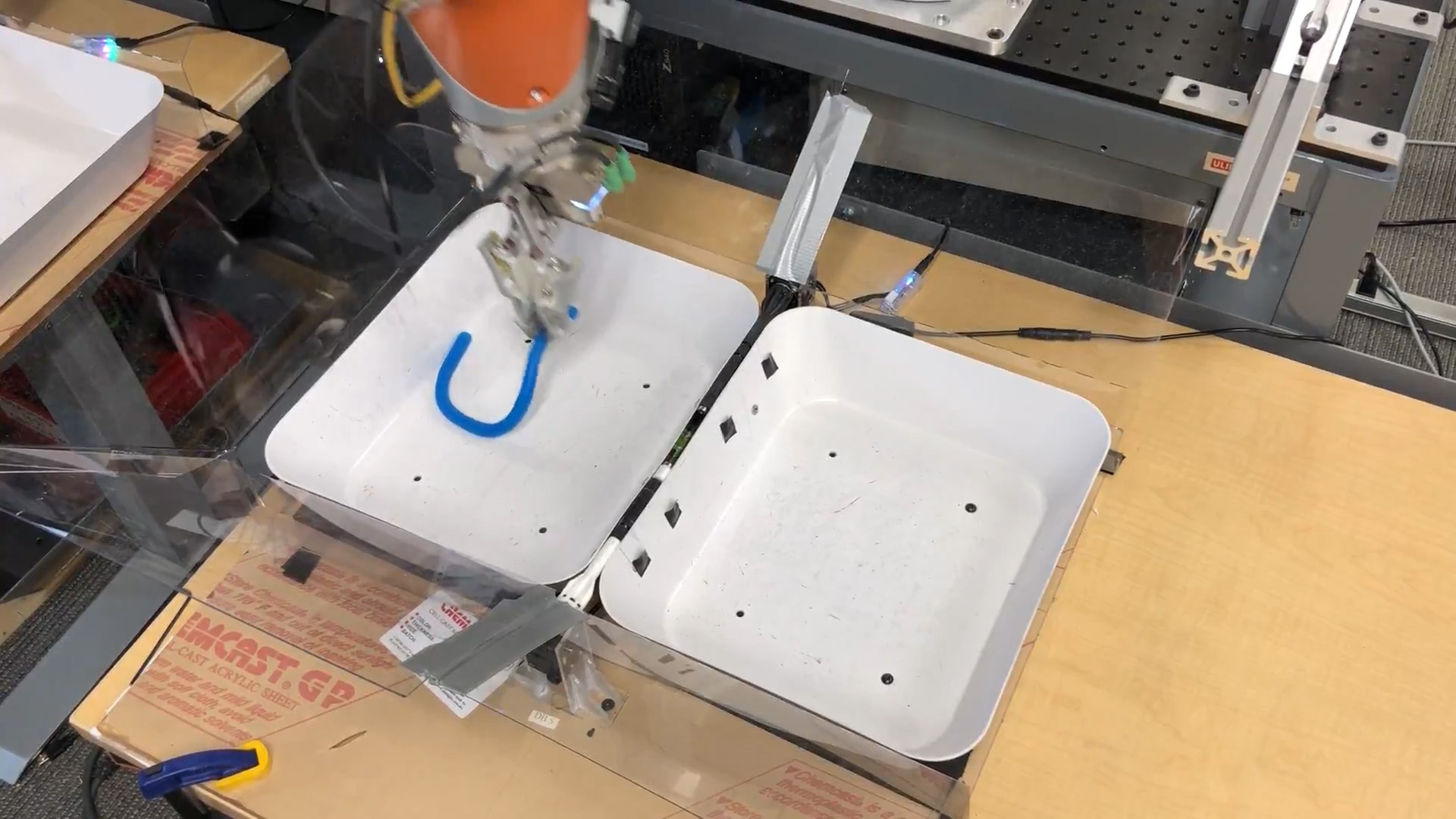}
    \includegraphics[trim=11cm 1.2cm 30cm 0cm, clip=true, width=0.173\columnwidth]{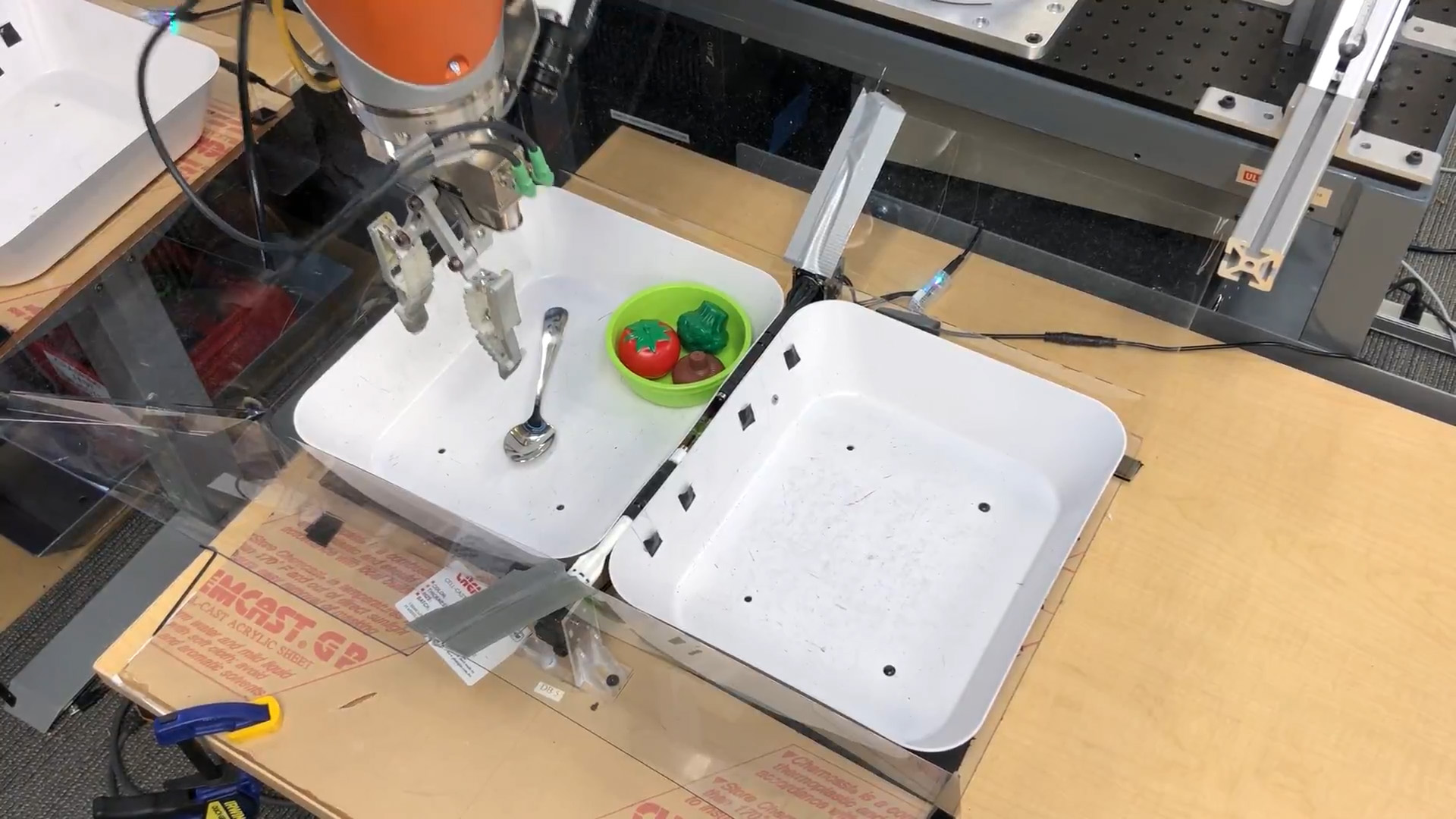}
    \includegraphics[trim=11cm 1.4cm 31cm 0cm, clip=true, width=0.168\columnwidth]{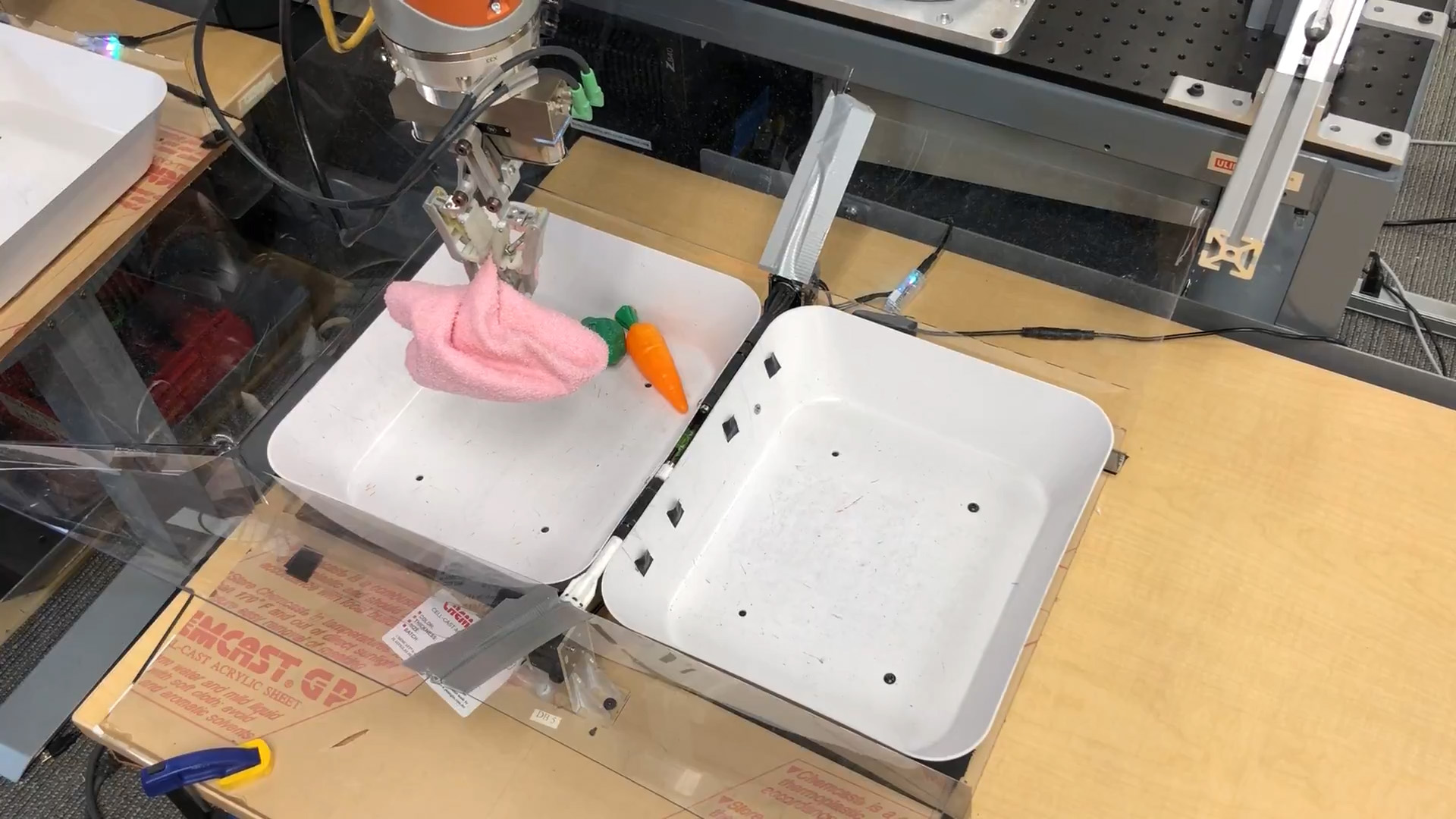}
    \vspace{-21pt}
    \caption{We propose a system for learning goal-reaching behaviors from offline datasets of real-world image-based experience. Our method enables a real-world robotic system to accomplish a wide range of visually indicated manipulation skills and learn rich representations that can  help  with  downstream  tasks.}
\label{fig:title}
\vspace{-10pt}
\end{figure}

\section{Introduction}
\label{sec:intro}

General-purpose robots will need large repertoires of skills. While reinforcement learning (RL) provides a way to learn such skills automatically, learning each skill individually can be prohibitive, particularly when task rewards must be programmed by hand and data must be collected anew for each task. 
Other fields of machine learning, such as natural language processing, have seen impressive progress from applying general-purpose training objectives on large and diverse datasets for either pre-training a general model that can be further fine-tuned on task-specific objectives~\cite{bert} or training a model that can be directly used for zero-shot or few-shot generalization~\cite{gpt3}.
Reusing data in robotic control therefore seems like an appealing prospect, but raises a number of questions:
How can we obtain a general-purpose training objective in robotics? How can we train diverse skills so that they are represented by a single model? How can we  employ this model to perform zero-shot generalization or solve downstream tasks?

We study these questions in the setting where the agent must learn entirely from \emph{offline} data, without hand-specified rewards or online interaction. This is particularly compelling for a general-purpose robotic system, which might already have data of past tasks that have been attempted (successfully or otherwise). Since the goal is to utilize past data from prior tasks to acquire transferable knowledge for \emph{new} tasks, we must determine what type of knowledge such data actually provides.
One answer offered in prior work is to use past experience to learn a functional understanding of the environment, as represented by a predictive model~\citep{deisenroth2011pilco,finn2017deep,bauza2017probabilistic}.
In this paper, we take a different perspective: instead of learning predictive models, which must predict future world observations in all of its complexity, we will learn goal-conditioned policies via offline, model-free RL. These policies must learn to reach any possible goal state from any current state, which as we show in this work, provides an effective general-purpose training objective that synthesizes varied tasks automatically, by considering every possible state reached in the data as a potential task.

While model-free training avoids the problem of predicting images, in the offline setting it carries its own challenges, such as distributional shift~\citep{laroche2017safe,kumar2019stabilizing}. 
Although prior works have sought to mitigate this challenge~\citep{lange2012batch,laroche2017safe,qtopt,kumar2019stabilizing,cql}, the goal-conditioned setting presents additional issues, since the offline data was not necessarily collected by a policy \emph{attempting} to reach any single goal. 
In this case, we must learn entirely from ``positive'' examples~\cite{xu2019positive, zolna2020offline}, which we show experimentally is challenging for current RL methods. 
We propose a strategy 
to provide synthetic ``negative'' labels that stabilize training by prescribing low Q-values for unseen actions, which, as indicated in prior work~\citep{kumar2019stabilizing}, also mitigates distributional shift in the offline setting. We further propose to extend this approach to learn how to reach goals that are \emph{more} temporally extended than the longest trajectory in the data, ``stitching'' together parts of the dataset into longer skills through \emph{goal chaining}. We call our framework \textit{Actionable Models} to emphasize the model-level granularity of offline data reuse and the ability to directly obtain an \textit{actionable} robotic policy.

Finally, having trained such \textit{Actionable Models} using a self-supervised goal-conditioned RL objective, we can utilize them to solve downstream tasks in three different ways: we can directly command goals to attain zero-shot generalization to new visually indicated tasks, we can use the goal-conditioned RL objective as an auxiliary loss for joint training, and we can fine-tune an Actionable Model with a task-specific reward.
\vspace{-1pt}
\section{Related Work}
\vspace{-2pt}
\textbf{Goal-conditioned RL.}
The aim of this work is to learn a goal-conditioned policy, a problem studied in many prior works~\citep{Kaelbling93, sutton2011horde, uvf, her, pong2018temporal, EysenbachGLS20, kadian2020sim2real, qtopt_goal}. Like prior work, we employ hindsight relabeling, reusing each trajectory to learn how to reach many goals.
Analogous to prior work~\citep{Kaelbling93, pong2018temporal}, we interpret the predictions from our Q-function as indicating the likelihood of reaching a particular goal in the future. We note that this is different from predicting an expected future state or representation~\citep{dayan1993improving, barreto2016successor, janz2018successor}.
Importantly, our setting differs from prior work in this area because we assume an \emph{offline} setting where the agent cannot interact with the environment to collect new data. Additionally, our system learns entirely from image observations for both the state and goal, without any manually specified similarity measure or reward function, in contrast to prior works that use distances~\citep{pong2018temporal} or $\epsilon$-ball threshold functions~\citep{her}.

\textbf{Goal-conditioned supervised learning.}
Prior methods for goal-conditioned supervised learning~\citep{oh2018self, ding2019goal, ghosh2019learning, sun2019policy,LynchKXKTLS19}  have a similar motivation: learning ``useful'' information for control without access to rewards. While these methods employ demonstrations~\citep{LynchKXKTLS19} or online data collection~\citep{ghosh2019learning}, our approach requires neither, and makes use of dynamic programming (Q-learning) to ``stitch'' together multiple trajectories to reach goals from states without ever having seen a complete trajectory from the state to that goal~\citep{sutton1988learning, greydanus2019the, singh2020cog}.

\textbf{Techniques from RL.}
Our method builds upon a number of tools from the RL literature. 
To handle offline Q-learning, we develop a technique similar to  Conservative Q-Learning (CQL)~\citep{cql}, which regularizes the Q-function on out-of-distribution actions. To handle large datasets and complex real-world tasks, we build on the QT-Opt algorithm~\citep{qtopt}, using the same training system and target value computation. Similar to~\cite{qtopt_goal}, we extend QT-Opt to the goal-conditioned setting. We further combine it with conservative strategies to enable learning in an offline setting.

\textbf{\emph{Explicit} dynamics models.}
Learning general behavior from offline datasets has been studied in the context of model-based RL, which learns explicit forward models~\citep{deisenroth2011pilco, watter2015embed, williams2015model, chua2018deep, buckman2018sample, wang2019exploring, kaiser2019model, hafner2019learning, nagabandi2020deep}.
The problem of learning dynamics models is a special case of a more general problem of auto-regressive time-series prediction (e.g., video prediction~\citep{mathieu2015video, denton2018stochastic, kumar2019videoflow}).
Our work is related in that we can learn our model from an offline dataset without any online interaction or reward function.
These dynamics models offer different capabilities than a goal-conditioned Q-function. Whereas dynamics models answer the question ``What does the future look like?", goal-conditioned Q-functions answer the opposite: ``What is the probability the future looks like \emph{this}?" For reaching a particular goal, the goal-conditioned Q-function directly provides for the optimal action. To select actions, most model-based methods require an additional cost function; our method has no such requirement.
\vspace{-1pt}
\section{Preliminaries}
\vspace{-1pt}

Let $\mathcal{M} = (\mathcal{S}, \mathcal{A}, P, R, p_0, \gamma, T)$ define a Markov decision process (MDP), where $\mathcal{S}$ and $\mathcal{A}$ are state and action spaces, $P: \mathcal{S} \times \mathcal{A} \times \mathcal{S} \rightarrow \mathbb{R}_{+}$ is a state-transition probability function, $R: \mathcal{S} \times \mathcal{A} \rightarrow \mathbb{R}$ is a reward function, $p_0: \mathcal{S} \rightarrow \mathbb{R}_{+}$ is an initial state distribution, $\gamma$ is a discount factor, and $T$ is the task horizon. We use $\tau = (s_0, a_0, \dots, s_T, a_T)$ to denote a trajectory of states and actions and $R(\tau) = \sum_{t=0}^T \gamma^t R(s_t,a_t)$ to denote the trajectory reward.
Reinforcement learning methods find a policy $\pi(a | s)$ that maximizes the expected discounted reward over trajectories induced by the policy: 
$\mathbb{E}_{\pi}[R(\tau)]$ where $ s_0\sim p_0, s_{t+1}\sim P(s_{t+1} | s_t, a_t)$ and $a_t\sim \pi(a_t | s_t)$.
Instead of manually defining a set of reward functions, we will define one sparse reward function per goal:
\begin{equation*}
R(\tau, g) = R(s_T, a_T, g) = \mathds{1} [s_T=g], g \in \mathcal{G}. \label{eq:reward}
\end{equation*}
The set $\mathcal{G}$ is a space of goals, which in our work we consider to be the space of goal images. We assume that the episode terminates upon reaching the specified goal, so the maximum total reward is 1.
We use temporal-difference (TD) learning to maximize the expected return~\cite{Kaelbling93,dayan1993improving}, yielding a goal-conditioned Q-function, which describes the probability of reaching the goal state $g$ at time $t = T$:
\vspace{-4pt}
\begin{align*}
    Q^\pi(s_t, a_t, g) &= \mathbb{E}_{\pi}\left[\sum_t \gamma^t R(s_t, a_t, g) \right] \\
    &= P^\pi(s_T = g | s_t, a_t).
\vspace{-1pt}
\end{align*}
We then obtain a goal-reaching policy by acting greedily with respect to the goal-conditioned Q-function: $\pi(a | s, g) = \argmax_a Q(s,a,g)$.
\section{Actionable Models}

In this section, we present our method, which we call \textit{Actionable Models}, for learning goal-conditioned skills from offline datasets.
Our method is based on goal-conditioned Q-functions with hindsight relabeling, for which offline data presents a particularly challenging setting. Hindsight relabeling only generates examples for actions that are needed to reach a goal, and does not provide evidence for which actions are sub-optimal or do not lead to a desired goal, which might result in over-estimation of their Q-values~\citep{schroecker2020universal}.
As we show in our experiments, without proper regularization, standard hindsight relabeling generally fails to learn effective Q-functions in an offline setting.
We present a scalable way to regularize Q-values in the offline setting by taking a \textit{conservative} approach for unseen actions, which was recently shown to help with distributional shift in offline learning~\citep{cql, singh2020cog}.
After introducing a framework for training goal-conditioned Q-functions offline, we further extend our method to enable \textit{goal chaining}, a way to reach goals across multiple episodes in the dataset, which enables long-horizon goal-reaching tasks.

\subsection{Goal-conditioned offline Q-learning}
\label{sec:goal_offline}
\begin{figure}[t]
    \centering
    \vspace{4pt}
    \includegraphics[trim=0cm 0cm 0cm 0cm, clip=true, width=0.57\columnwidth]{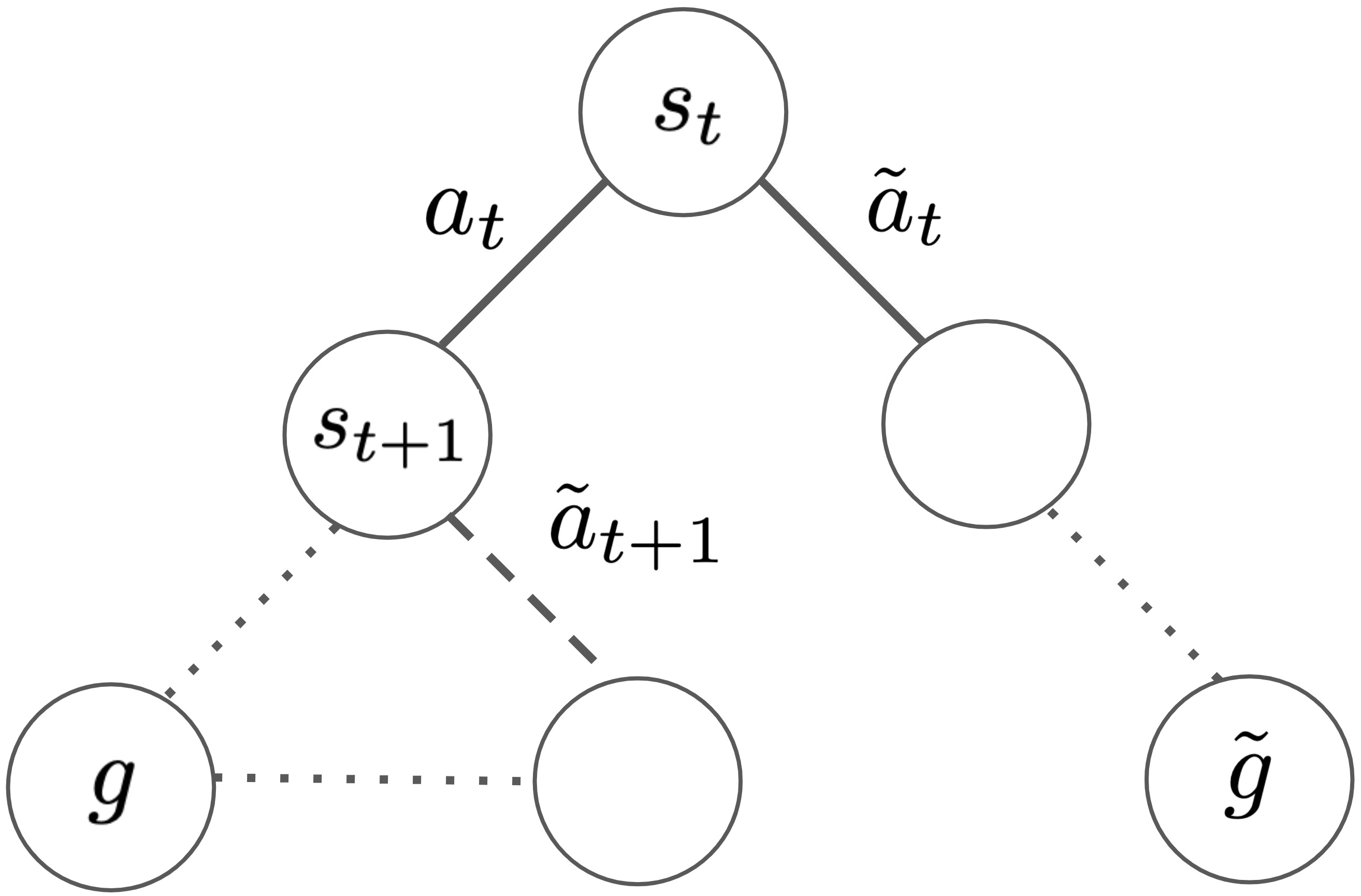} 
    \vspace{-10pt}
    \caption{Two scenarios that can occur when deviating from the original action on the way to the goal $g$: without recovery ($\tilde{a}_t$) and with recovery ($\tilde{a}_{t+1}$).}
    \label{fig:neg_actions}
    \vspace{-6pt}
\end{figure}
Given a dataset $\mathcal{D}$ of previously collected trajectories $\tau \sim \mathcal{D}$, we aim to learn to reach all goals in the dataset using a goal-conditioned policy  $\pi(a \mid s, g) = \argmax_{a} Q(s,a,g)$. 
Each trajectory $\tau$ provides evidence of which states are reachable from which other states in the trajectory. More formally,
each sub-sequence $\tau_{0:i} = (s_0, a_0,...,s_i)$ with $i \in (0,...,T)$ is a \textit{positive} example of reaching the goal $g = s_i$. Thus, we label each such sub-sequence with the success-reward $\mathds{1}$ given the goal $g=s_i$:
\[
R(\tau_{0:i}, g=s_i) = \mathds{1} [s_i=s_i] = 1.
\]
Such relabeling alone is not enough to train a well-conditioned Q-function with respect to the action $a_t$. In particular, we do not know which actions result in not reaching the desired goal. 

Let $ \tilde{\mathcal{A}}(s, g)$ denote a set of unseen actions, for which we do not have evidence of reaching the goal $g$ from state $s$ in the dataset $\mathcal{D}$, and $\tilde{a} \sim p_{\tilde{\mathcal{A}}} (\tilde{a} | s, g)$ some probability distribution with the support on this set, which we will describe below. Furthermore, let $\tilde{\mathcal{G}}(g)$ be a set of goals different from $g$ and $p_{\tilde{\mathcal{G}}}(g)$ a distribution with the support on this set.
Fig.~\ref{fig:neg_actions} demonstrates two possible scenarios when deviating from  original actions seen in the dataset: either the policy does not recover and lands in another state $\tilde{g} \sim p_{\tilde{\mathcal{G}}}(g)$ (e.g., $\tilde{a}_t$ in the figure), or it recovers from the deviation and reaches the desired goal $g$ (e.g., $\tilde{a}_{t+1}$ in the figure).

We follow a conservative approach that avoids over-estimation of Q-values for unseen actions. We assume there is no recovery unless there are trajectories in the dataset $\mathcal{D}$ that demonstrate this recovery, which would be taken into account through labeling them with the success-reward as described above. This amounts to assuming that any deviation $\tilde{a} \sim p_{\tilde{\mathcal{A}}} (\tilde{a} | s, g)$ leads to some other goal $\tilde{g} \sim p_{\tilde{\mathcal{G}}}(g)$:
\begin{align*}
\mathbb{E}_{\tilde{a}_t\sim p_{\tilde{\mathcal{A}}}} \left[P^\pi(s_T \neq g|s_t, \tilde{a}_t)\right] = 1.
\end{align*}
With high-dimensional goals like images, the set of all other goals $\tilde{\mathcal{G}}(g)$ becomes intractable. We can circumvent dealing with $\tilde{\mathcal{G}}(g)$ by the following transformation:
\begin{align*}
\mathbb{E}&_{\tilde{a}_t\sim p_{\tilde{\mathcal{A}}}} \left[P^\pi(s_T \neq g|s_t, \tilde{a}_t) \right]
\\&= \mathbb{E}_{\tilde{a}_t\sim p_{\tilde{\mathcal{A}}}}\left[1 - P^\pi(s_T = g|s_t, \tilde{a}_t)\right] \\
&= \mathbb{E}_{\tilde{a}_t\sim p_{\tilde{\mathcal{A}}}}\left[1 - Q^\pi(s_t, \tilde{a}_t, g)\right] = 1 \\
&\Rightarrow \mathbb{E}_{\tilde{a}_t\sim p_{\tilde{\mathcal{A}}}}\left[Q^\pi(s_t, \tilde{a}_t, g)\right] = 0.
\end{align*}
This means, we aim at minimizing Q-values for unseen actions $\tilde{a}_t \sim p_{\tilde{\mathcal{A}}} (\tilde{a}_t | s_t, g)$, which we denote as \textit{negative} actions as opposed to \textit{positive} actions $a_t$ obtained through relabeling from the dataset. We find that sampling contrastive $\tilde{a}_t$ close to the decision boundary of the Q-function -- i.e., unseen actions that have high Q-values -- to be the most effective implementation of this objective. Therefore, we define $p_{\tilde{\mathcal{A}}} (\tilde{a} | s, g) \sim \frac{1}{Z}\exp \left (Q(s_t, \tilde{a}_t, g) \right)$, sampling negative actions according to the soft-max distribution over their Q-values.
This has an interesting connection to conservative Q-learning (CQL)~\cite{cql}, which although having a different motivation of improving out-of-distribution behavior of Q-learning, arrives at a similar objective of minimizing Q-values for unseen actions. In particular, CQL shows that by sampling  $\tilde{a}_t\sim \exp(Q^\pi(s_t, \tilde{a}_t, g))$, it is possible to learn a Q-function that lower bounds the true Q-function on unseen actions.

\subsection{Goal chaining}
\label{sec:goal_chaining}

Using the procedure in the previous subsection, we can train a policy to reach goals within all sub-sequences $\tau_{0:i}$ in the dataset $\tau \sim \mathcal{D}$,
but the policy would only learn to reach goals reachable within a single trajectory $\tau$ in the dataset.
However, parts of tasks can often be spread out across multiple trajectories, such as when accomplishing the goal of one trajectory is a pre-requisite for beginning to attempt the goal of another trajectory.

Ideally, in such situations we would like to be able to specify the final goal without manually specifying intermediate goals.
The dynamic programming nature of Q-learning is able to concatenate or \textit{chain} trajectories in state space, but further modifications are needed in order to enable chaining in \textit{goal} space. 
Although similar techniques have been proposed in previous works~\cite{her, NairPDBLL18, lin2019reinforcement} for online learning, we develop a way to perform goal chaining in an offline~setting.

\begin{figure}[t]
    \centering
    \includegraphics[trim=0cm 0cm 0cm 0cm, clip=true, width=0.72\columnwidth]{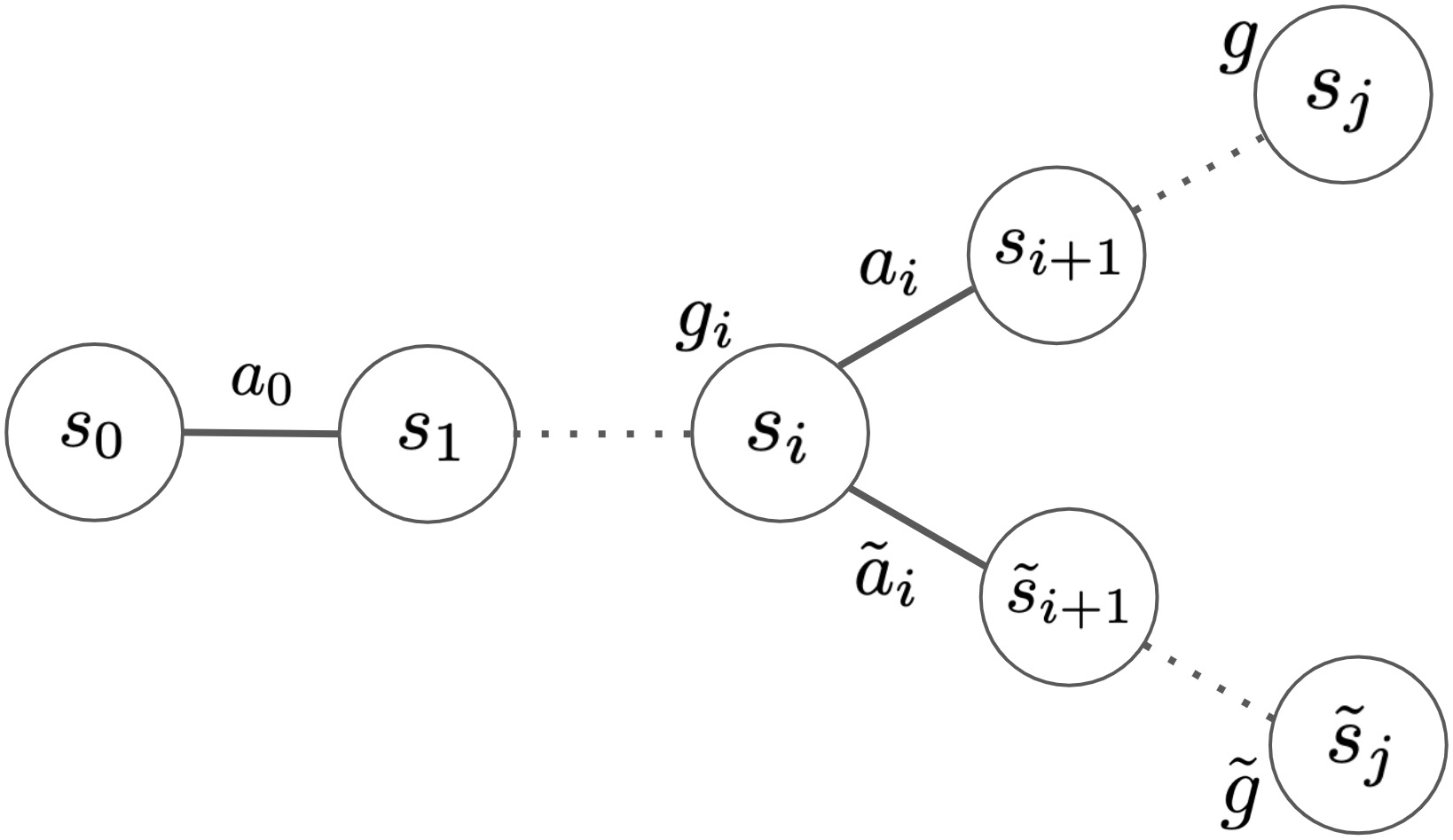} 
    \vspace{-12pt}
    \caption{Illustrating goal chaining between two trajectories: $\tau_{0:i}$ and $\tau_{i:j}$ through the chaining point $s_i$.}
    \label{fig:goal_chain}
    \vspace{-9pt}
\end{figure}

To enable long-horizon tasks through goal chaining, we propose a simple modification to our goal-conditioned Q-learning method. Given a sequence $\tau_{0:i}$, instead of limiting the goal to be within the sequence (i.e., $g \sim (s_0,...,s_i)$), we redefine it to be any state observed in the dataset (i.e., $g \sim \mathcal{D}$).
If $g=s_i$ (e.g. when the goal is the final state of the sub-sequence) then similarly as before, we label such trajectories with $R(\tau_{0:i},s_i)=1$.
Otherwise, since we do not know whether $g\neq s_i$ can be reached from the states within $\tau_{0:i}$, instead of assigning a constant reward, we set the reward of the final transition to be its Q-value, such that $R(s_i, a_i, g) = Q^\pi(s_i, a_i, g)$:
\[
R(\tau_{0:i}, g)  = 
\begin{cases}
  1, & \mbox{if } s_i=g \\
  Q^\pi(s_i,a_i,g), & \mbox{otherwise.}
\end{cases}
\]
This procedure follows the intuition that if there is evidence in the dataset that $g$ is reachable from $(s_i, a_i)$, it will eventually propagate into the Q-values of $\tau_{0:i}$. Fig.~\ref{fig:goal_chain} illustrates this intuition. Without loss of generality, assume that there exist a sequence $\tau_{i:j}$ that starts with $(s_i, a_i)$ and leads to $s_j=g$. Then there will be a constant reward $R(s_{j}, a_{j}, g) = 1$, which will propagate to $Q^\pi(s_i, a_i, g)$ and eventually to  $Q^\pi(s_{0:i-1}, a_{0:i-1}, g)$. In this case, we call $Q^\pi(s_i, a_i, g)$ a \textit{chaining point} between two trajectories $\tau_{0:i}$ and $\tau_{i:j}$. 
It should be noted that without introducing such chaining points, e.g. through additional random goal labeling as done in this work, Q-learning alone might not be able to connect these trajectories as they will have distinct goals from within their corresponding sequences. 
To handle the situation when there is no pathway from $s_i$ to $g$, we can still apply the conservative regularization technique described in the previous subsection. In particular, as we minimize Q-values for actions $\tilde{a}_i\sim\exp (Q^\pi(s_i, \tilde{a}_i, g))$, without an evidence for reachability of $g$ our method will assume that the policy ends up in a different goal $\tilde{g}\sim\tilde{\mathcal{G}}$ and eventually also minimize $Q^\pi(s_{0:i-1}, a_{0:i-1}, g)$. 

Interestingly, this kind of chaining works across more than two episodes. In fact, it is only limited by the long-horizon propagation of values through the Bellman recursion and by the discount factor $\gamma$, which attenuates the constant reward from the last trajectory in the chain. More generally, this procedure connects to previous works on learning goal-distance functions, such as~\citet{Kaelbling93}, where the objective is to learn distances between all states. Furthermore, we can generalize the goal sampling procedure by defining a goal sampling function $g \sim G(\mathcal{D}, s_i, a_i)$, which can further be adjusted, e.g. by skewing the sampling distribution towards goals with high Q-values similar to~\cite{EysenbachGLS20} to reduce the number of unreachable goals. It should also be noted that in the case when we use function approximators for learning Q-functions, the chaining points does not have to be exactly the same in both trajectories in order to propagate useful reachability information. As we show in the ablation experiments in Section~\ref{sec:goal_chaining_exp}, the goal chaining procedure significantly improves the capabilities of our learned policy for accomplishing longer-horizon tasks with high-dimensional image goals.

\vspace{-1pt}
\subsection{Method Summary}
\vspace{-2pt}
\label{sec:method_summary}
We now summarize our complete method, which we call \textit{Actionable Models}. Our method takes as input a dataset of trajectories $\mathcal{D} = \{\tau\}$, and outputs a goal-conditioned Q-function, $Q_\theta(s, a, g)$. Note that acting greedily w.r.t. this goal-conditioned Q-function provides us with a goal-conditioned policy: $\pi(a | s, g) = \arg\max_a Q(s, a, g)$. We train this Q-function by minimizing the following loss:
\begin{align}
    \mathcal{L}_g(\theta) \!= & \min_\theta \mathbb{E}_{(s_t,a_t,s_{t+1},g)\sim\mathcal{D}}[(Q_\theta(s_t, a_t, g) \! - \!y(s_{t+1}, g))^2 \nonumber \\
    &+ \mathbb{E}_{\tilde{a}\sim\exp(Q_\theta)}[(Q_\theta(s, \tilde{a}, g) - 0)^2]], \label{eq:goal-objective}
\end{align}
where the TD-target $y$ is given by:
\begin{equation*}
    y(s_{t+1}, g) = \begin{cases}1 & \text{if }s_{t+1} = g \\ \gamma \mathbb{E}_{a\sim\pi}[Q_\theta(s_{t+1}, a, g)] & \text{otherwise.} \end{cases}
\end{equation*}
The first part of the loss $\mathcal{L}_g(\theta)$ increases Q-values for reachable goals and the second part regularizes the Q-function. It should be noted that when optimizing with gradient descent, the loss does not propagate through the sampling of action negatives $\tilde{a}\sim\exp(Q_\theta)$.

\vspace{-2pt}
\section{Applications of Actionable Models}
\vspace{-3pt}
In this section, we present applications of our framework, including learning general goal-reaching policies and using the goal-reaching objective to learn rich representations that can be used in downstream RL tasks, e.g. through pre-training or an auxiliary optimization objective.

\vspace{-1pt}
\subsection{Goal reaching}
\vspace{-2pt}
Our framework can be applied for learning goal reaching skills from previously collected offline datasets. Algorithm~\ref{algo:am} outlines the example extraction and Q-target computation components.
This method can be integrated into any Q-learning method with an experience replay buffer~\cite{mnih2015humanlevel}. 

The \textsc{\footnotesize{ExtractExamples}} module samples a trajectory from the dataset and randomly cuts it to produce a sub-sequence $\tau_{0:i}$. We  relabel this sub-sequence with $R(\tau_{0:i}, s_i) = 1$ for reaching its final state as in Section~\ref{sec:goal_offline}, and with $R(\tau_{0:i}, g_{rand}) = Q_\theta(s_i, a_i, g_{rand})$ for some goal $g_{rand}\sim\mathcal{D}$ to enable goal chaining as in Section~\ref{sec:goal_chaining}. All transitions from the relabeled trajectories are added to the replay buffer.

In \textsc{\footnotesize{ComputeQTargets}}, we first sample transitions with relabeled rewards $(s_t, a_t, g, s_{t+1}, R(s_t, a_t, g))$ from the replay buffer and apply the Bellman equation to compute Q-targets for these transitions. Then, we sample a negative action $\tilde{a}_t \sim \exp (Q_\theta(s_t, \tilde{a}_t, g))$ by first uniformly sampling actions from our action space and then reweighting them by their $\exp (Q_\theta)$-values.
When sampling negative actions, we filter out actions that are too close to the seen action $a_t$.

\begin{figure}[t]
\vspace{-8pt}
\begin{algorithm}[H]
\begin{algorithmic}[1]
\small
\FUNCTION{\textsc{\footnotesize{ExtractExamples}}}
\STATE{$\tau \gets \mathcal{D}$: Sample a trajectory from the dataset.}
\STATE{$g_{rand} \gets \mathcal{D}$: Sample a random goal from the dataset.}
\STATE{$\tau_{0:i} \gets$ Randomly cut the trajectory with $i\in\{1,T\}$.}
\STATE{$R(\tau_{0:i}, s_i) = 1$: Label reaching final state with 1.}
\STATE{$R(\tau_{0:i}, g_{rand}) = Q_\theta(s_i, a_i, g_{rand})$: \\ Label reaching $g_{rand}$ with Q-value at the final state.}
\STATE{Add transitions from relabeled trajectories to replay buffer.}
\ENDFUNCTION
\FUNCTION{\textsc{\footnotesize{ComputeQTargets}}}
\STATE{$(s_t, a_t, g, s_{t+1}, R(s_t, a_t, g)) \gets$ \\ Sample a transition from replay buffer.}
\STATE{$Q_{target}(s_t, a_t, g) \gets $ \\$R(s_t, a_t, g) + \max_a Q(s_{t+1}, a, g)$}
\STATE{$\tilde{a}_t \sim \exp (Q_\theta(s_t, \tilde{a}_t, g))$: Sample a negative action.}
\STATE{$Q_{target}(s_t, \tilde{a}_t, g) \gets 0$: \\ Set the target for the negative action to 0.}
\ENDFUNCTION
\end{algorithmic}
\caption{Goal reaching with Actionable Models}
\label{algo:am}
\end{algorithm}
\vspace{-27pt}
\end{figure}

\vspace{-1pt}
\subsection{Pre-training}
\vspace{-2pt}
\label{sec:pretrain}
Besides using actionable models for goal reaching, we can also leverage their ability to learn functional representations of the world to accelerate acquisition of downstream tasks through conventional reward-driven online reinforcement learning. Given a large dataset $\mathcal{D}$ of previous experience, we pre-train a goal-conditioned Q-function $Q_\theta(s,a,g)$ using our offline method, and then further fine-tune it on a specific task reward.
As we will show in our experiments in Section~\ref{sec:exp_repr_learning}, pre-training on large and diverse datasets with actionable models leads to representations that significantly accelerate the acquisition of downstream tasks. 

\vspace{-1pt}
\subsection{Auxiliary objective}
\vspace{-2pt}
\label{sec:aux}
In addition to pre-training, we can also utilize our method to provide an auxiliary objective that can be used in parallel with conventional online RL to encourage learning of functional representations.
Given a small mix-in probability $\xi$ we can augment the task-specific objective $\mathcal{L}_{task}(\theta)$ with the regularized goal-reaching objective $\mathcal{L}_{g}(\theta)$ from Eq.~\ref{eq:goal-objective}:
\[
\mathcal{L}_{augmented}(\theta) = \mathcal{L}_{task}(\theta) + \xi \mathcal{L}_{g}(\theta).
\]
To optimize this joint objective, we add three kinds of data to our replay buffer. With probability $\xi$, we add relabeled trajectories using \textsc{\footnotesize{ExtractSamples}}, which will be used to optimize the $\mathcal{L}_g$ term. With probability $\xi$, we create a relabeled trajectory but replace the action with an unseen action and replace the reward with 0 (following \textsc{\footnotesize{ComputeQTargets}}).
With probability $(1 - 2\xi)$, we add the original trajectories (no relabeling) with the original rewards.
When training on original trajectories (no relabeling), we condition the Q-function on a goal image set to all zeros.
\section{Experiments}
\label{sec:experiments}
In our experiments, we aim to answer the following questions: 1) How does our method compare to previous methods for learning goal-conditioned policies from offline data, such as goal-conditioned behavioral cloning and standard Q-learning with hindsight relabeling? 2) Can our method learn diverse skills on real robots with high-dimensional camera images? 3) Does our proposed goal chaining technique facilitate learning to reach long-horizon goals? 4) Is goal reaching an effective pre-training step or a suitable auxiliary objective for accelerating learning of downstream reward-driven skills?

\subsection{Experimental Setup}
\begin{figure}[t]
    \centering
    \vspace{-2pt}
    \includegraphics[trim=0.4cm 0cm 0.5cm 0cm, clip=false, width=0.911\linewidth]{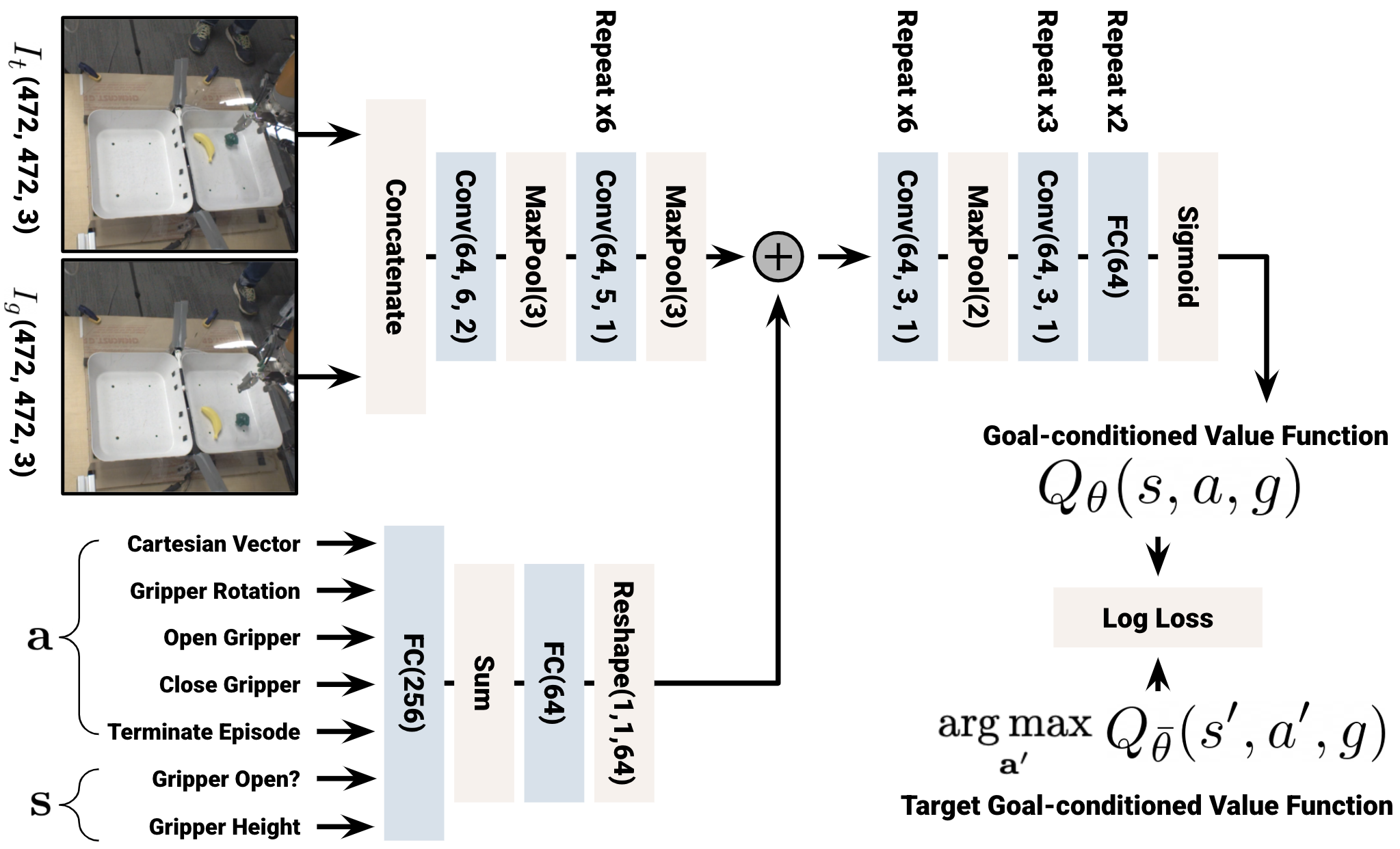}
    \vspace{-10pt}
    \caption{The Q-function neural network architecture follows the QT-Opt framework with an additional goal image input.}
    \label{fig:nn}
    \vspace{-10pt}
\end{figure}

We evaluate our method on a variety of robotic manipulation scenarios, with actions corresponding to Cartesian space control of the robot's end-effector in 4D space (3D position and azimuth angle) combined with discrete actions for opening/closing the gripper and terminating the episode.
In all experiments, our policy architecture follows the QT-Opt framework~\cite{qtopt}, with an additional Q-function input for the 472x472x3 goal image, which is concatenated with the current robot camera image before being fed into a convolutional tower. The full set of inputs and the network architecture can be found in Fig.~\ref{fig:nn}. We also use the cross-entropy loss to fit the Q-target values, as in the original QT-Opt framework. 
The initial variances of the $(x,y,z)$ end-effector positions and the azimuth angle during the CEM-optimization~\cite{cem} are set to (3cm, 3cm, 6cm, 0.16 rad), respectively. We run CEM for 2 iterations with 64 samples per iteration and 10\% elite percentile. More details on the implementation of QT-Opt can be found in prior work~\cite{qtopt}.

\begin{figure*}[!ht]
    \centering
\subfloat{
\includegraphics[trim=0cm 0cm 0cm 0cm, clip=true, width=0.8\linewidth]{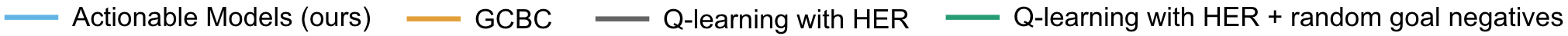}} \\ \vspace{-12pt}
\hspace*{-5pt}\setcounter{subfigure}{0}
    \subfloat[task1][Pick-and-place]{
\includegraphics[trim=0cm 0cm 0cm 0cm, clip=true, width=0.246\linewidth]{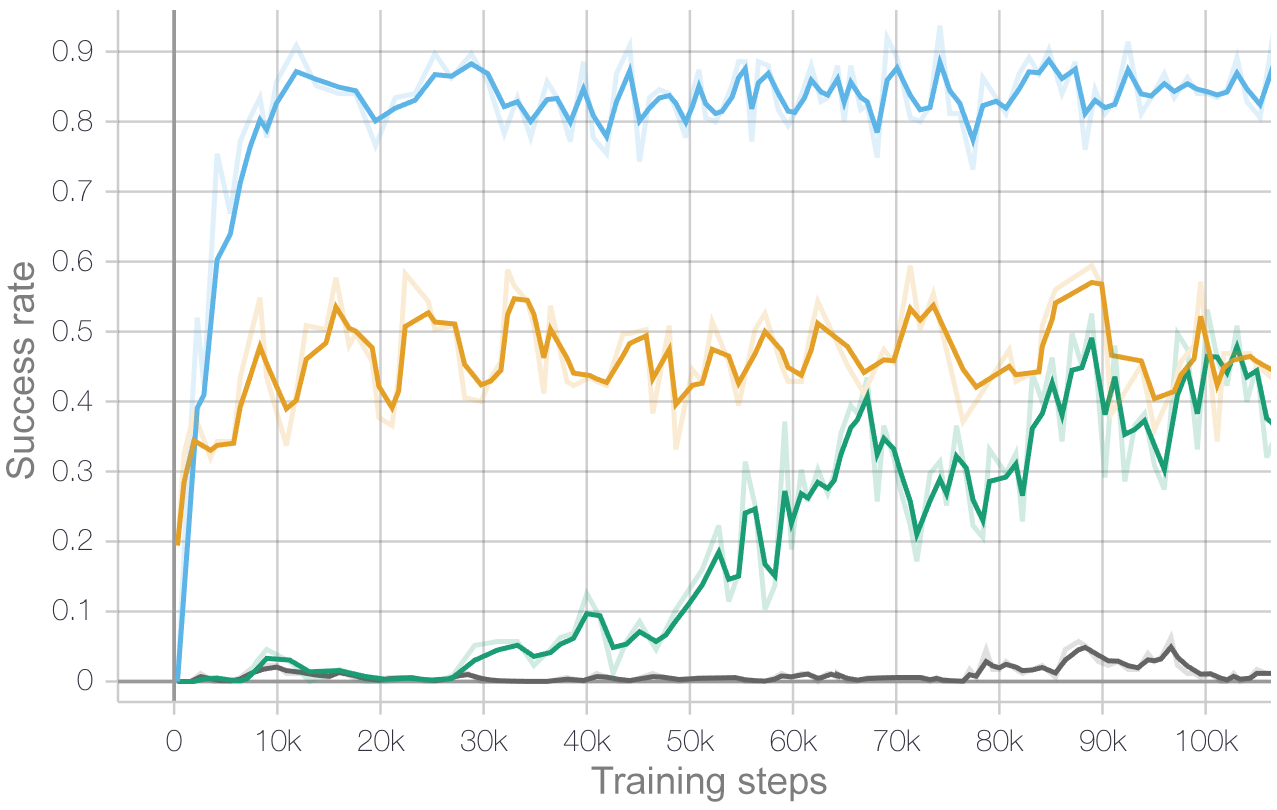}}
    \subfloat[task11][Stacking]{
\includegraphics[trim=0cm 0cm 0cm 0cm, clip=true, width=0.246\linewidth]{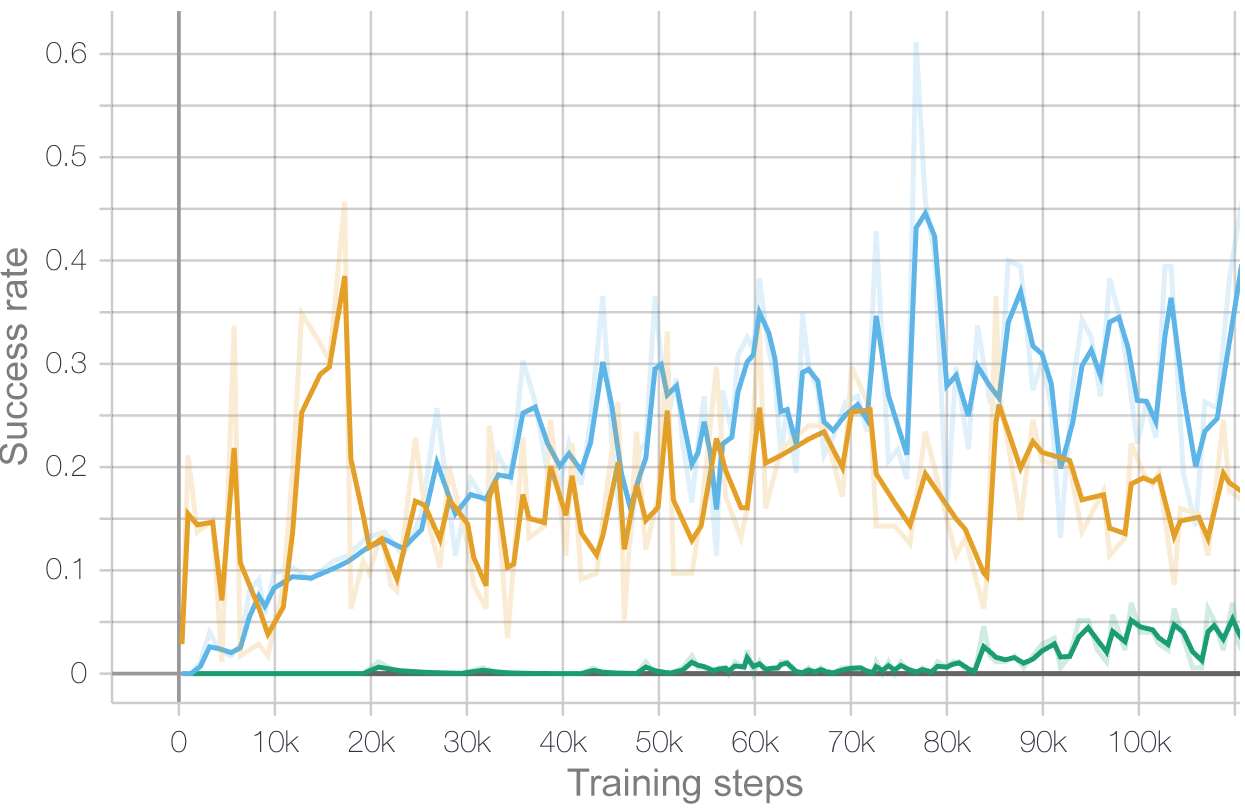}}
    \subfloat[task67][Fixture placing]{
\includegraphics[trim=0cm 0cm 0cm 0cm, clip=true, width=0.246\linewidth]{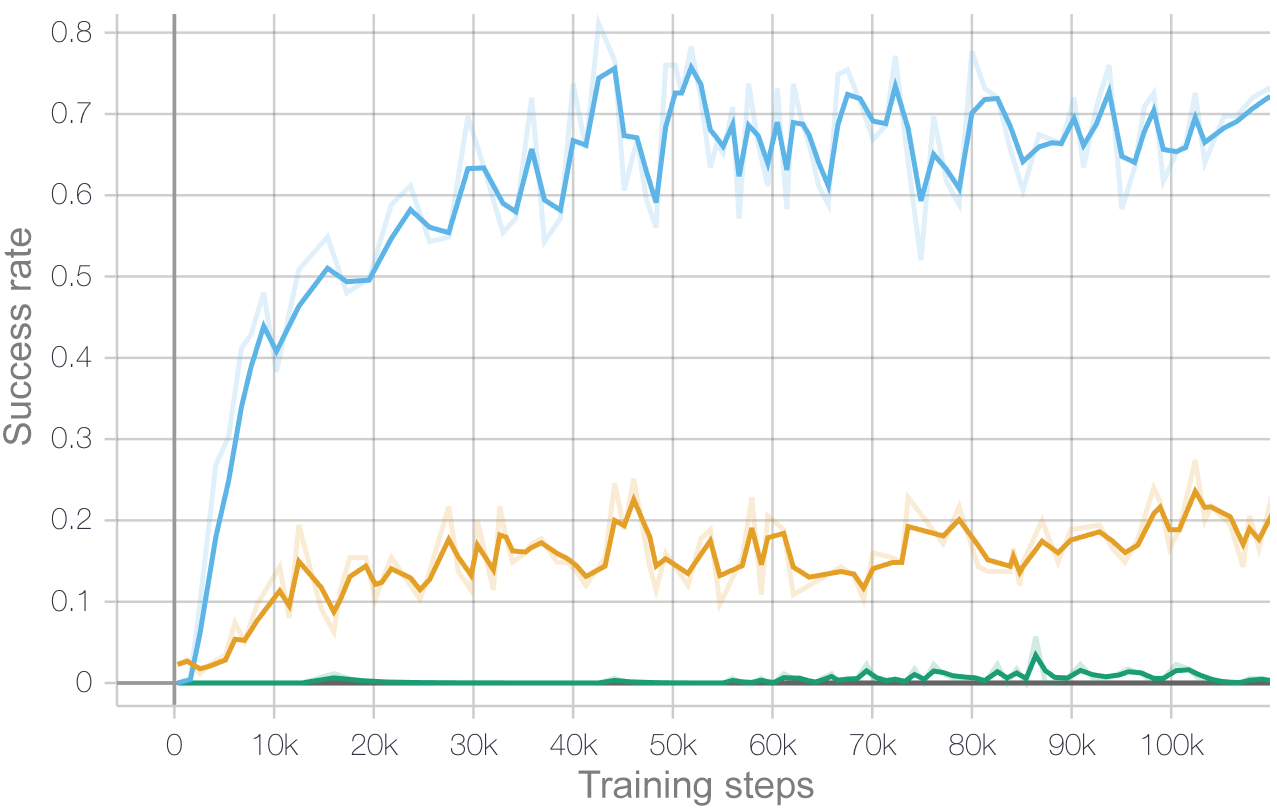}}
    \subfloat[task78][Food object grasping]{
\includegraphics[trim=0cm 0cm 0cm 0cm, clip=true, width=0.246\linewidth]{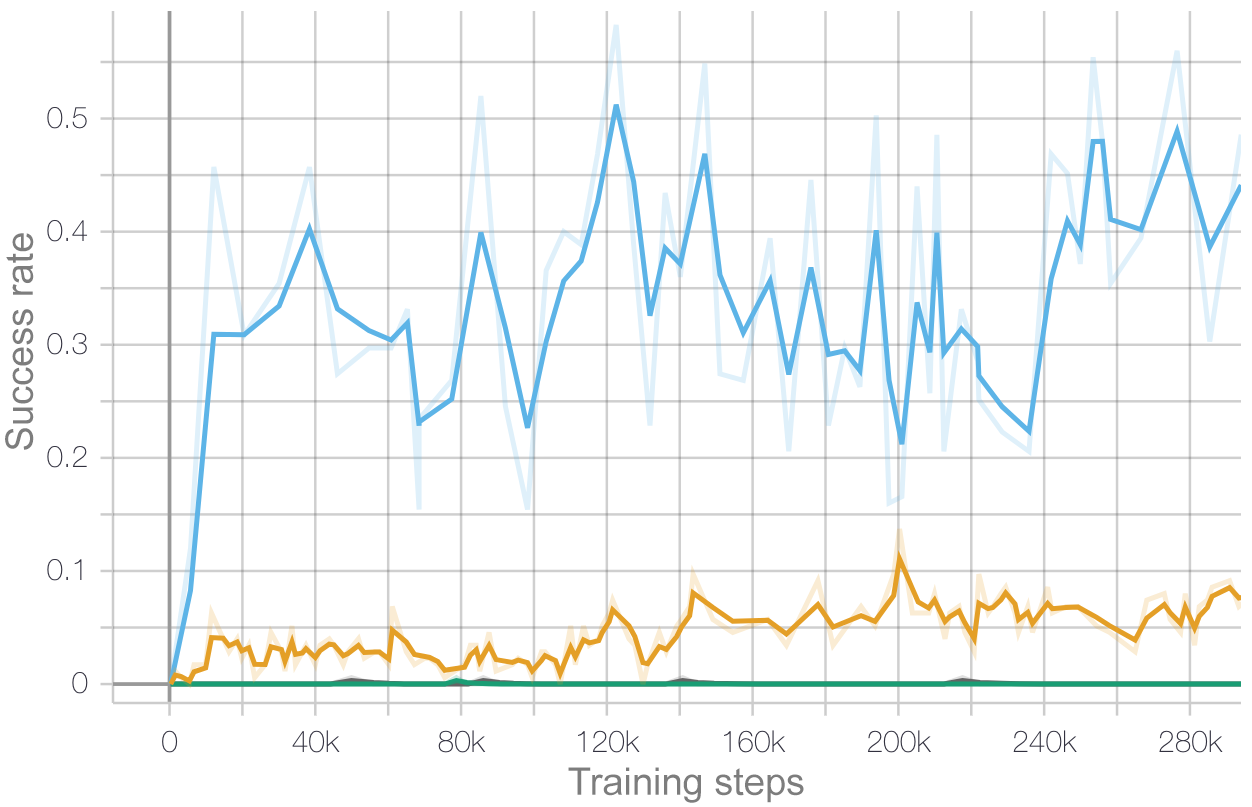}}
\vspace{-4pt}
\caption{Comparison of goal-conditioned policies trained from offline data in simulation.}
\label{fig:goal_conditioned_comp}
\vspace{-6pt}
\end{figure*}
\begin{figure}[t]
\vspace{-10pt}
    \centering
     \includegraphics[trim=6cm 3cm 0cm 0cm, clip=true, width=0.24\columnwidth]{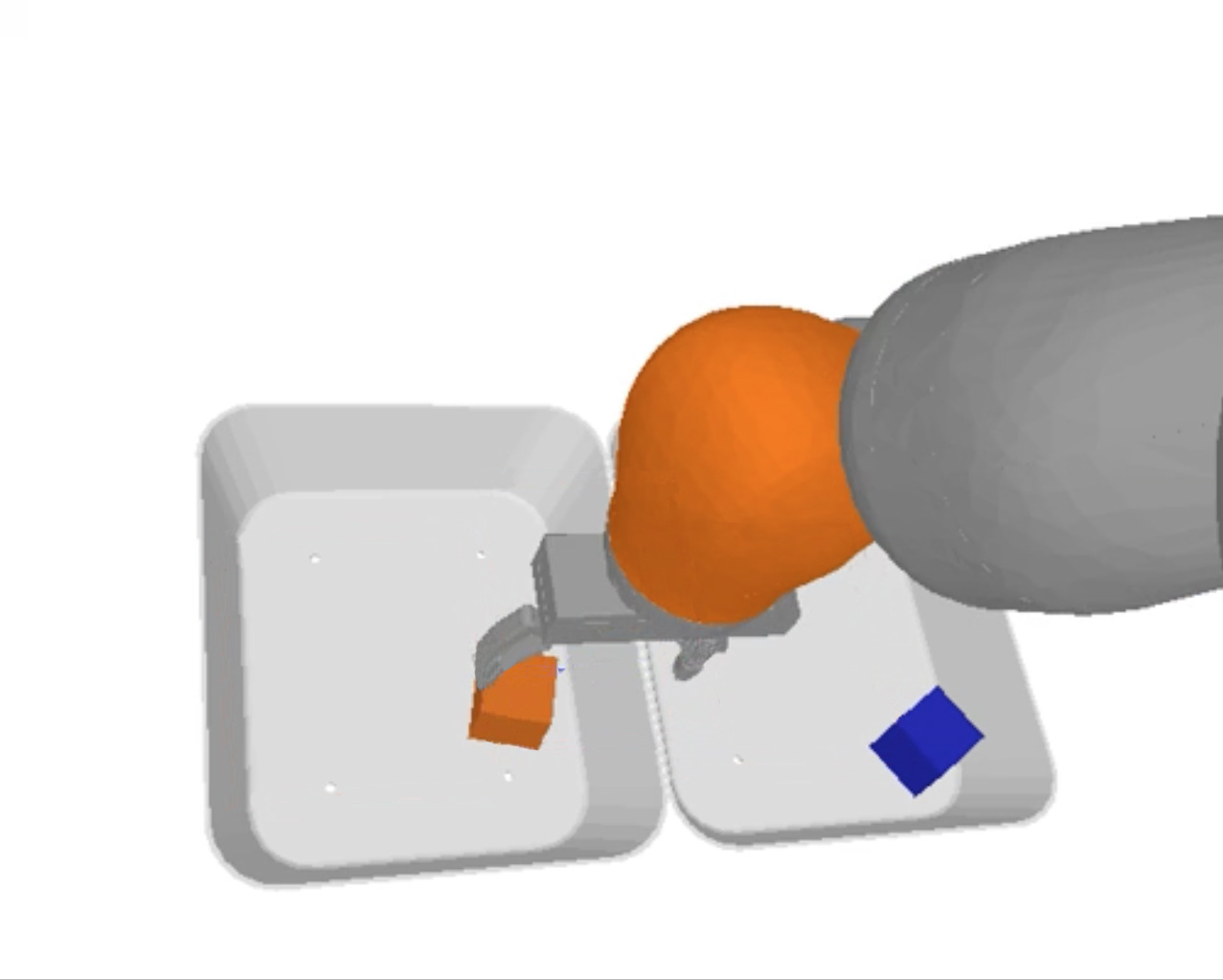}
    \includegraphics[trim=6cm 3cm 0cm 0cm, clip=true, width=0.24\columnwidth]{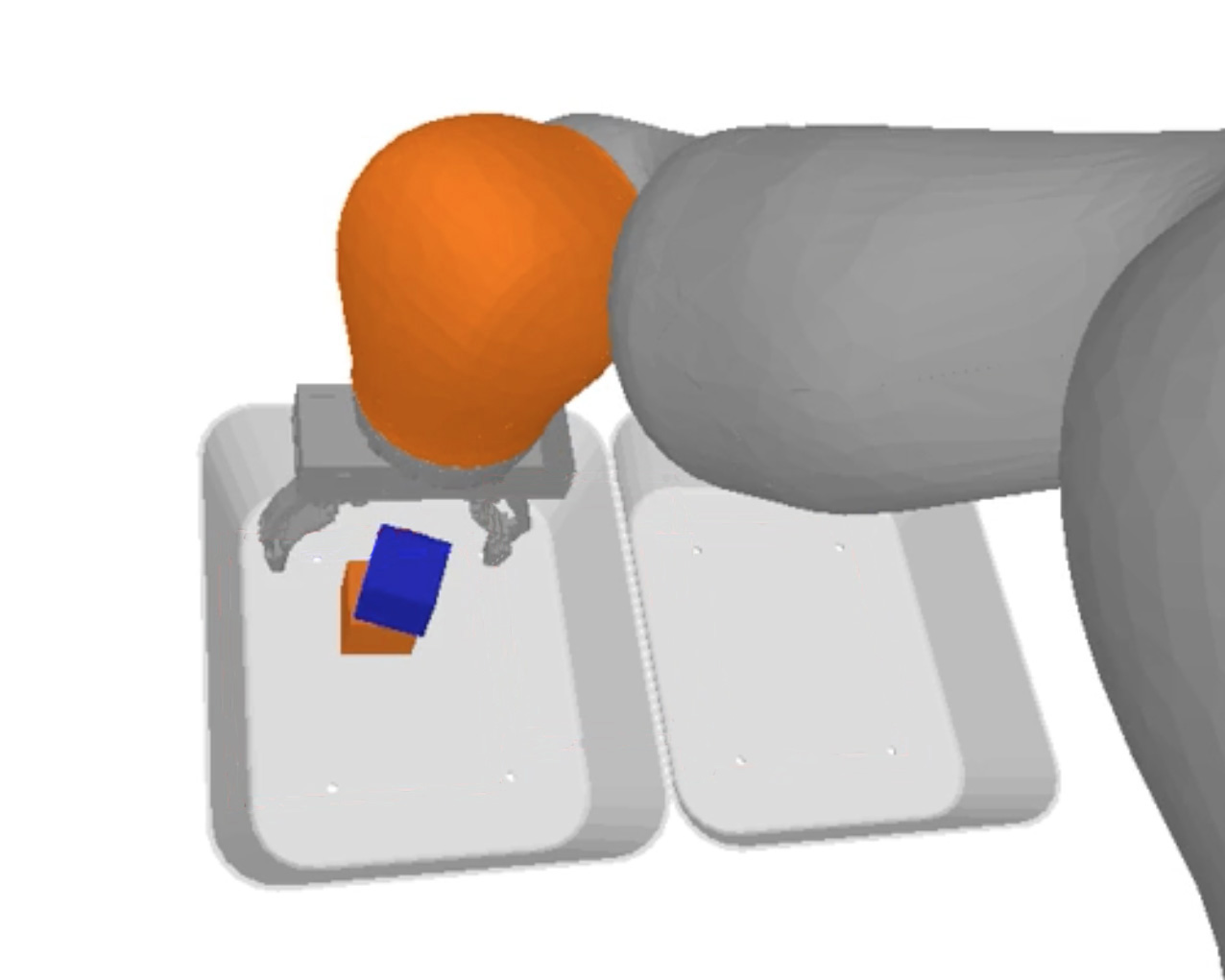}
    \includegraphics[trim=6cm 3cm 0cm 0cm, clip=true, width=0.24\columnwidth]{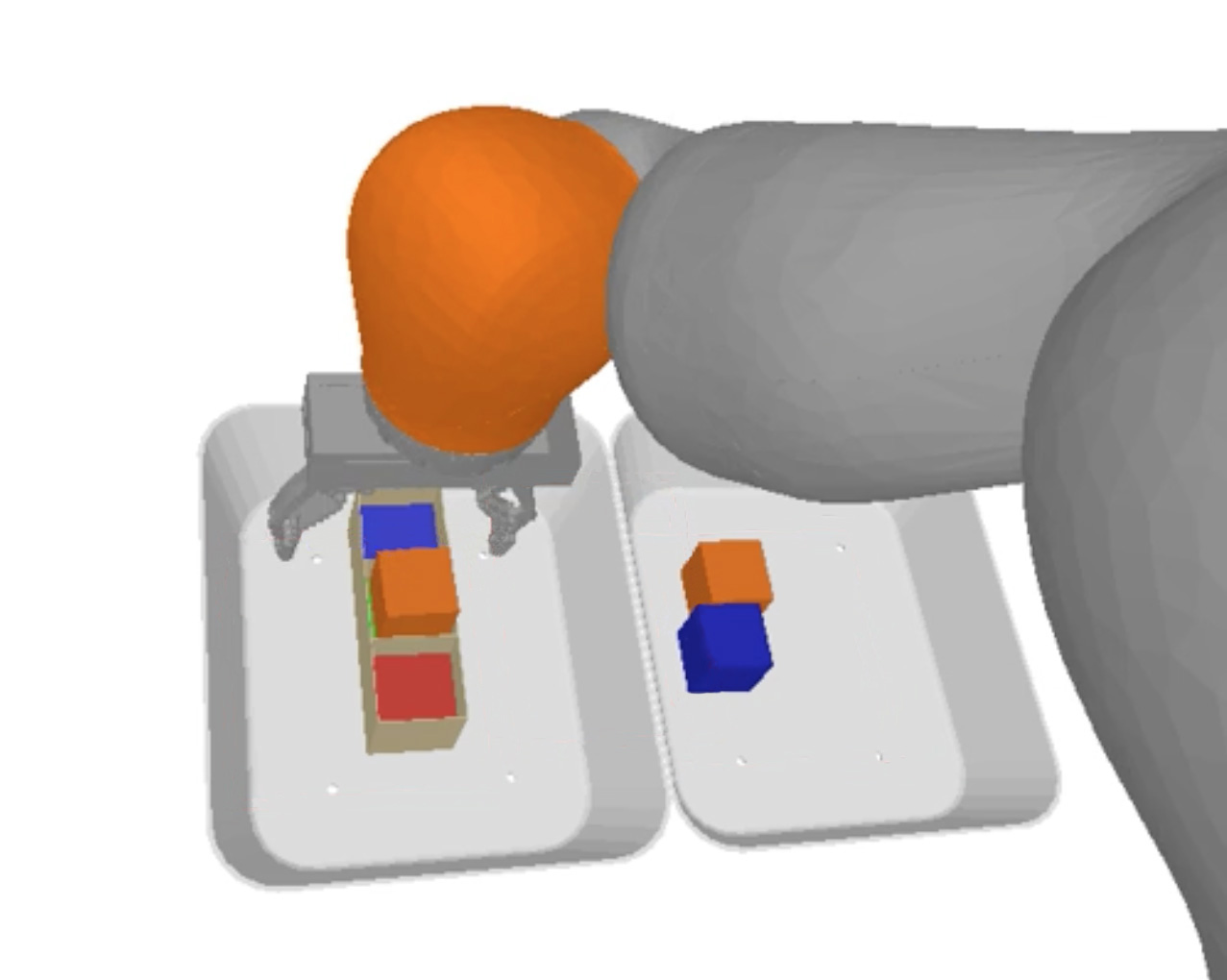}
    \includegraphics[trim=6cm 3cm 1cm 1cm, clip=true, width=0.24\columnwidth]{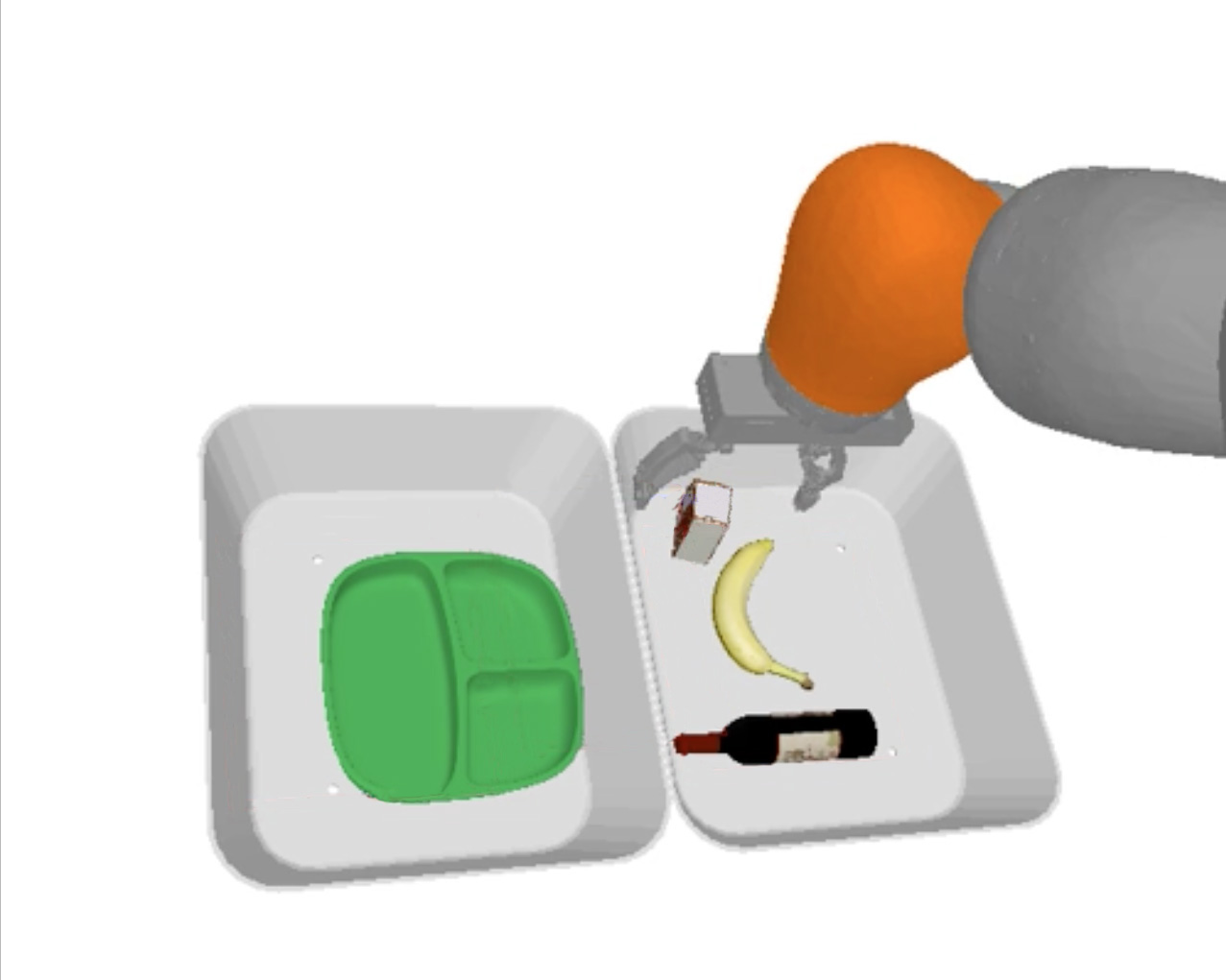}
\vspace{-8pt}
\caption{Simulated tasks (robot camera view): pick-and-place, stacking, fixture placing, food object grasping.}
\label{fig:sim_tasks}
\vspace{-5pt}
\end{figure}

\subsection{Training data}
Acquiring data with good state coverage and visual variety is important for successful application of offline reinforcement learning. In this work, we employ data coming from RL training traces for diverse tasks and environments in both simulation and the real world, which span the gamut from random exploration to fully-trained policies. In practical applications, we might expect Actionable Models to similarly be trained on diverse and heterogeneous datasets, which might combine demonstrations or play data~\cite{LynchKXKTLS19}, random exploration, and experience from other RL policies.

Our real robot training data comes from a wide range of reinforcement learning experiments conducted as part of another research project using the same platform~\cite{mtopt2021arxiv}.
Crucially, none of this data was generated by goal-conditioned policies. The data was produced by policies at various stages of training, from random initializations to near-optimal behaviors, trained for 12 different tasks (with conventional reward functions) using mostly a diverse set of food objects that can be seen in Fig.~\ref{fig:real_robot_train}. The success rates in the data for the different tasks ranged from 5\% to 55\%, and the tasks included various versions of grasping and pick-and-place skills. The data contains over 800,000  episodes, which altogether amounts to over 6 months of continuous robotic interaction on 7 KUKA robots. 

\vspace{-2.7pt}
\subsection{Simulated visual goal reaching experiments}
\vspace{-2.3pt}
\label{sec:sim_goal_cond}
In our simulated experiments, we generate a dataset of trajectories from RL runs for the four tasks depicted in Fig.~\ref{fig:sim_tasks}. We train expert policies for each task with their task-specific rewards using QT-Opt until convergence (all expert policies reach more than $90\%$ success rate at convergence). We then combine transitions from these training runs to train a single unified goal-image conditioned Q-function with our method.
To the best of our knowledge, prior methods for model-free goal-conditioned RL require online exploration, whereas our problem setting requires fully offline training. In the absence of clear points of comparison, we devise a number of baselines that we believe most closely reflect widely used prior methods, which we adapt to the offline setting:

\begin{figure*}[t]
    \captionsetup[subfigure]{justification=centering}
\subfloat{\includegraphics[trim=0.1cm 0cm 0cm 0cm, clip=true, width=0.999\linewidth]{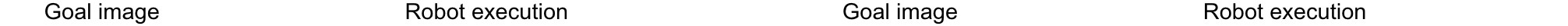}}\\\vspace{-21pt}

\subfloat{\includegraphics[trim=5cm 0cm 0cm 3.5cm, clip=true, width=0.1154\linewidth]{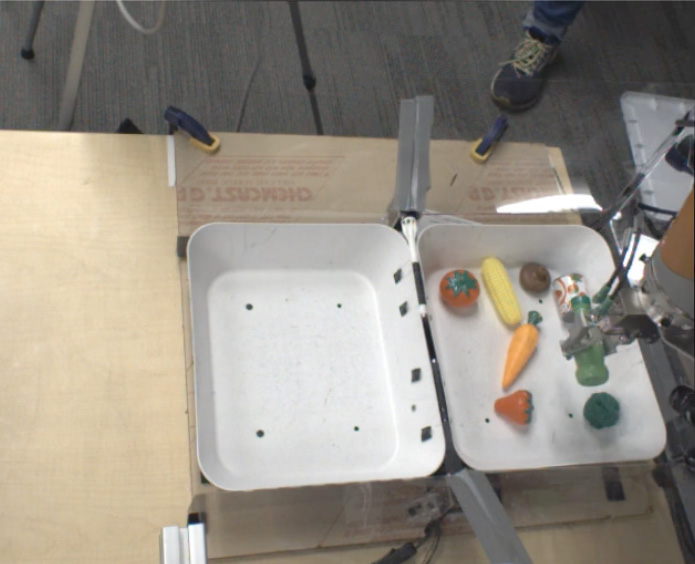}\hspace{2pt}\includegraphics[trim=0cm 3.8cm 0cm -0.1cm, clip=true,width=0.008\linewidth]{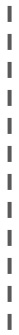}\hspace{2pt}\includegraphics[trim=5cm 0cm 0cm 3.5cm, clip=true, width=0.1154\linewidth]{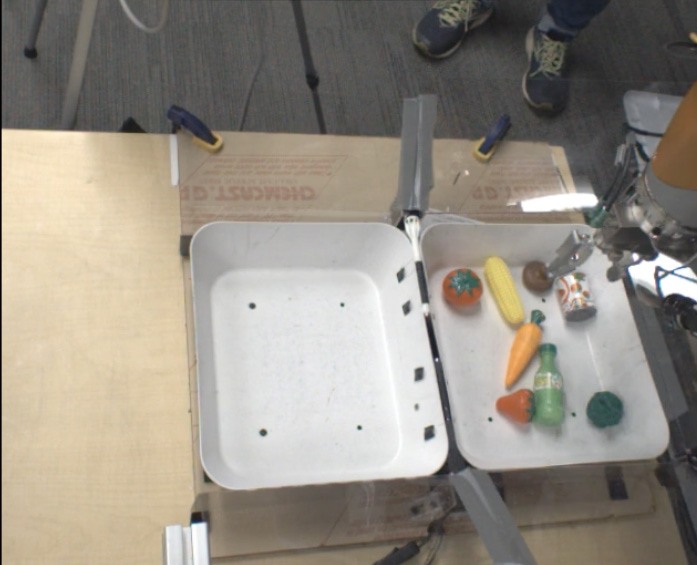} \includegraphics[trim=5cm 0cm 0cm 3.5cm, clip=true, width=0.1154\linewidth]{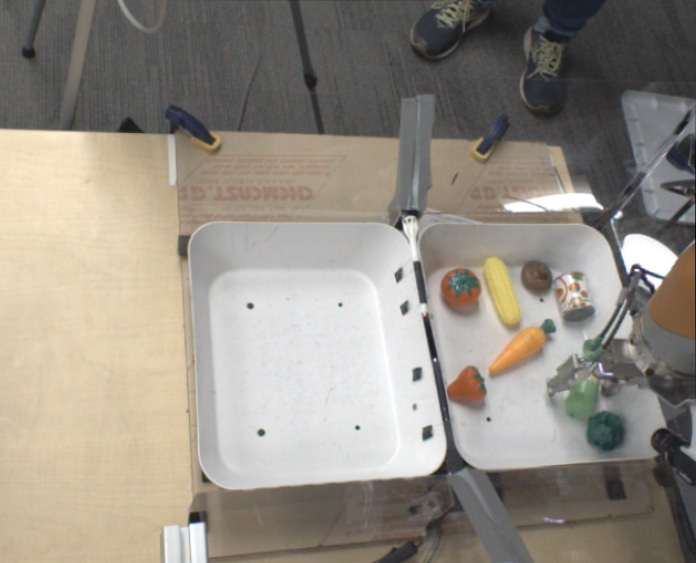} \includegraphics[trim=5cm 0cm 0cm 3.5cm, clip=true,width=0.1154\linewidth]{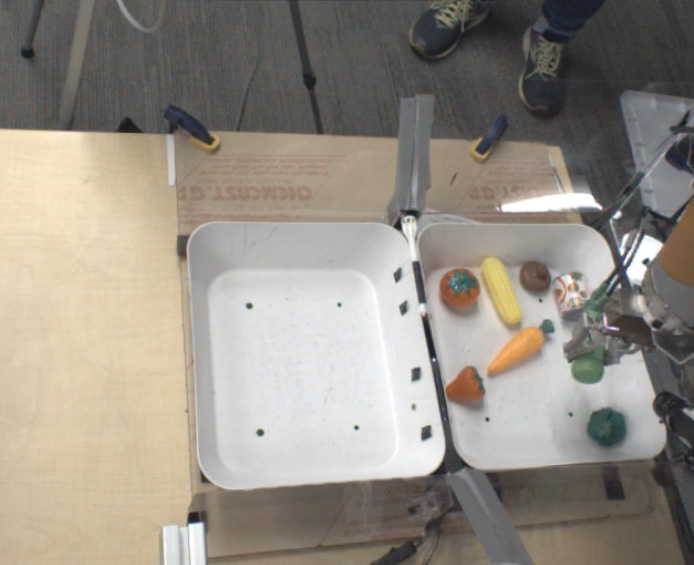}}\hspace{9pt}
\subfloat{\includegraphics[trim=5cm 0cm 0cm 3.5cm, clip=true, width=0.1154\linewidth]{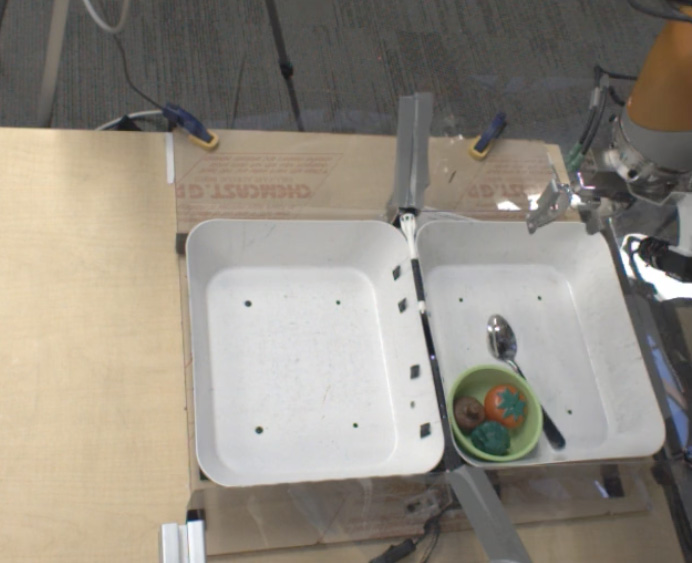}\hspace{2pt}\includegraphics[trim=0cm 3.8cm 0cm -0.1cm, clip=true,width=0.008\linewidth]{images/real_robot/real_camera_sep.jpg}\hspace{2pt}\includegraphics[trim=5cm 0cm 0cm 3.5cm, clip=true, width=0.1154\linewidth]{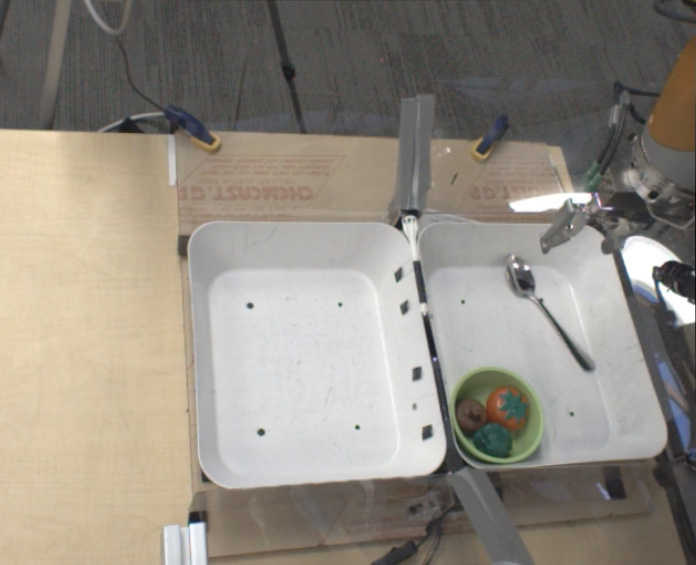} \includegraphics[trim=5cm 0cm 0cm 3.5cm, clip=true, width=0.1154\linewidth]{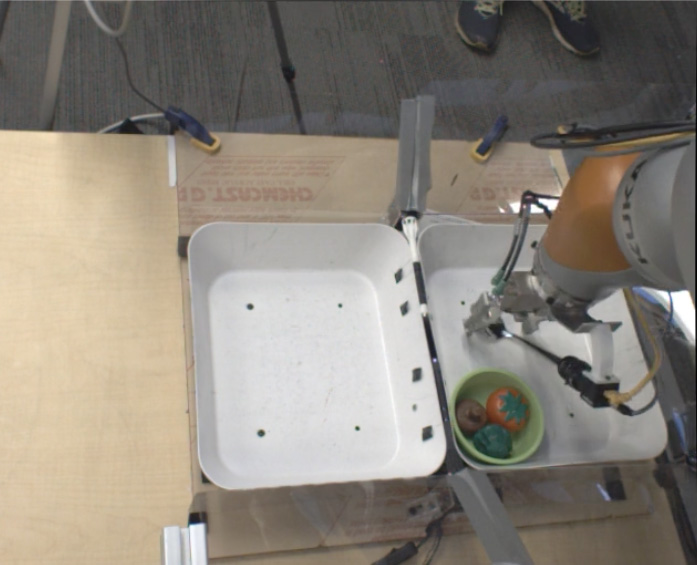} \includegraphics[trim=5cm 0cm 0cm 3.5cm, clip=true,width=0.1154\linewidth]{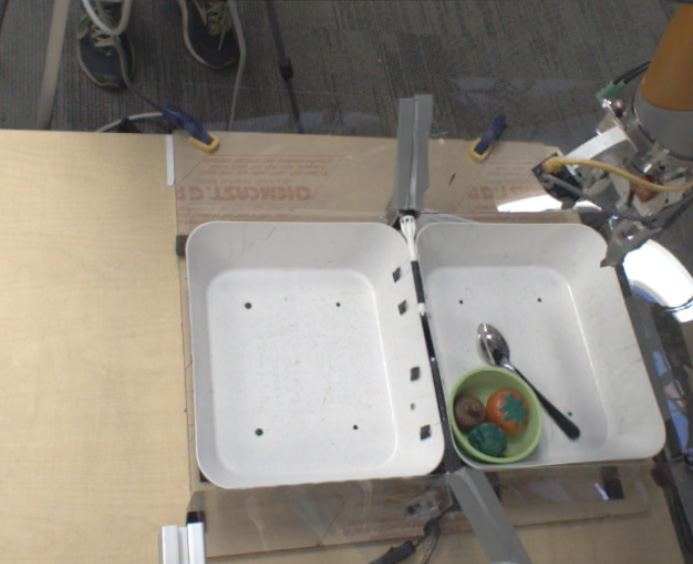}}\\\vspace{-21pt}

\subfloat{\includegraphics[trim=5cm 0cm 0cm 3.5cm, clip=true, width=0.1154\linewidth]{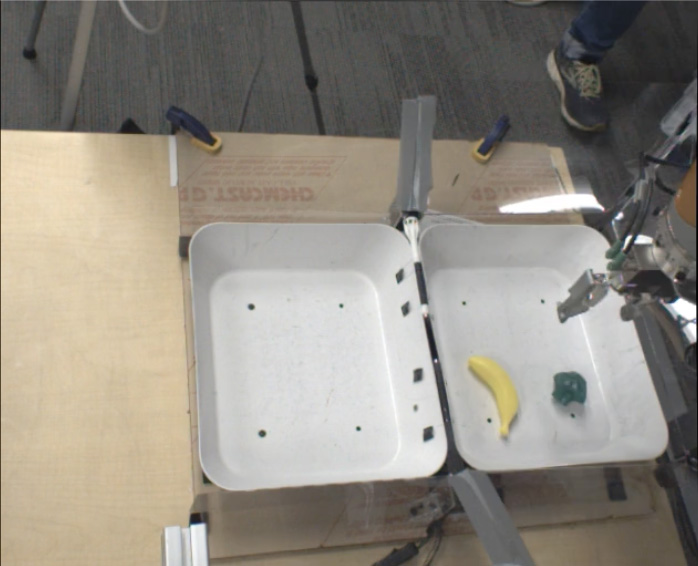}\hspace{2pt}\includegraphics[trim=0cm 3.8cm 0cm -0.1cm, clip=true,width=0.008\linewidth]{images/real_robot/real_camera_sep.jpg}\hspace{2pt}\includegraphics[trim=5cm 0cm 0cm 3.5cm, clip=true, width=0.1154\linewidth]{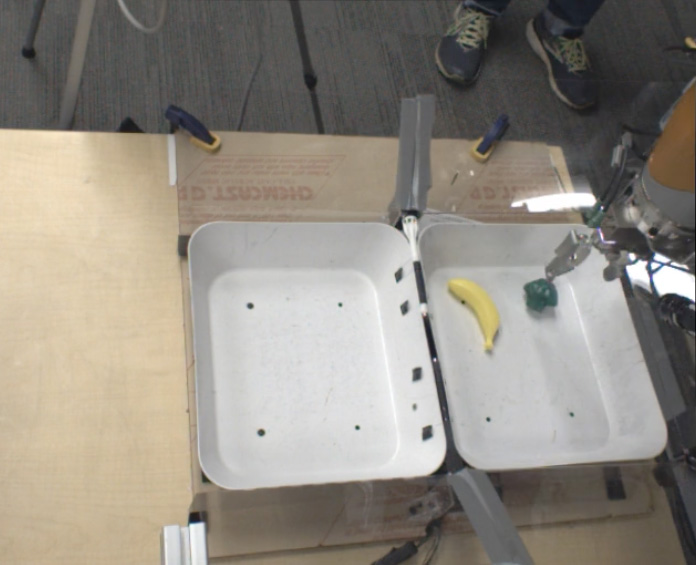} \includegraphics[trim=5cm 0cm 0cm 3.5cm, clip=true, width=0.1154\linewidth]{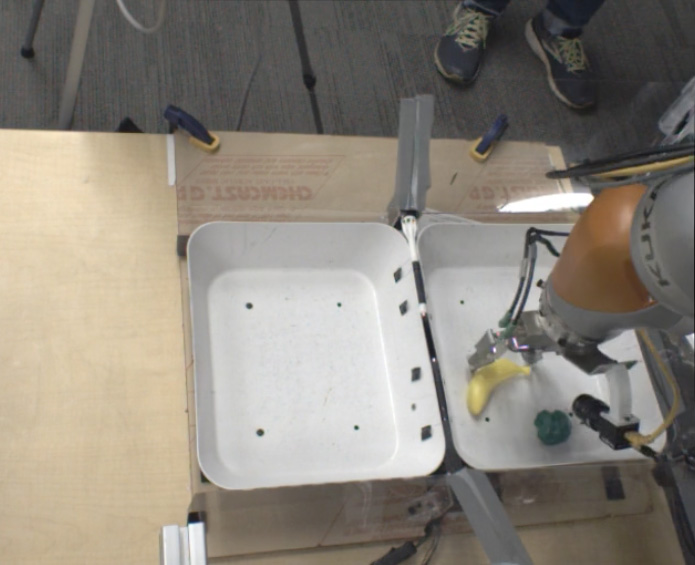} \includegraphics[trim=5cm 0cm 0cm 3.5cm, clip=true,width=0.1154\linewidth]{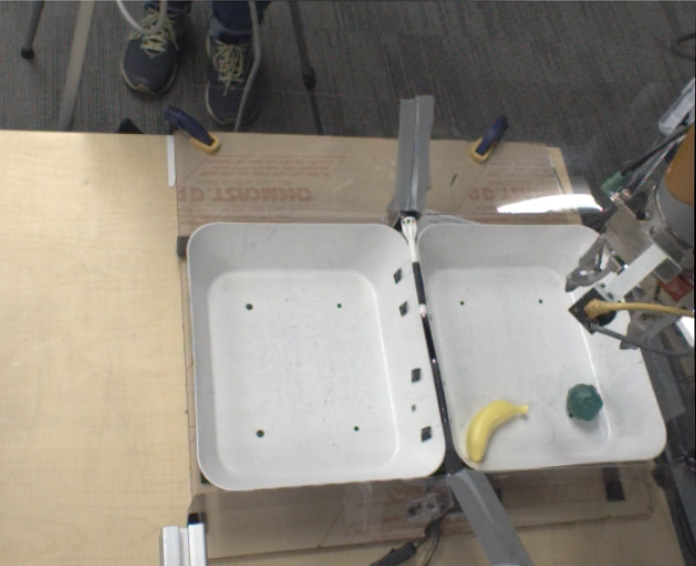}}\hspace{9pt}
\subfloat{\includegraphics[trim=5cm 0cm 0cm 3.5cm, clip=true, width=0.1154\linewidth]{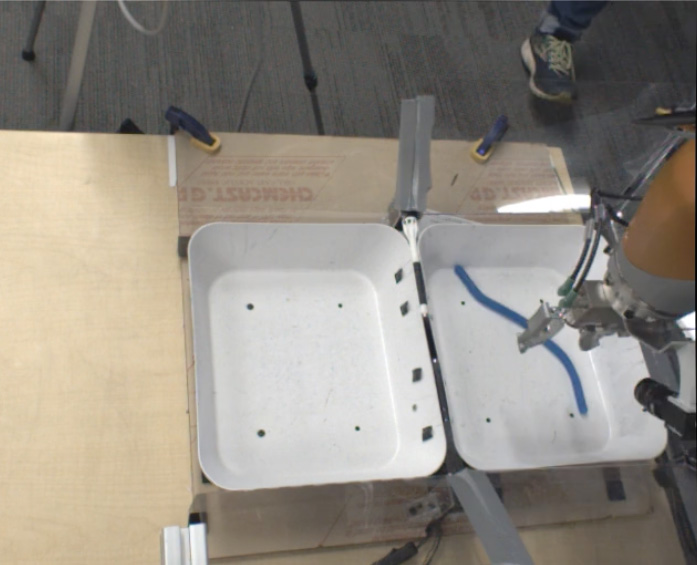}\hspace{2pt}\includegraphics[trim=0cm 3.8cm 0cm -0.1cm, clip=true,width=0.008\linewidth]{images/real_robot/real_camera_sep.jpg}\hspace{2pt}\includegraphics[trim=5cm 0cm 0cm 3.5cm, clip=true, width=0.1154\linewidth]{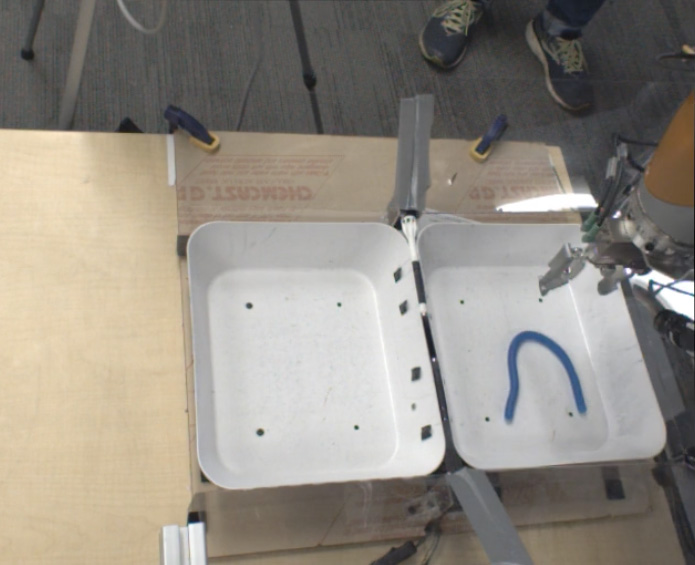} \includegraphics[trim=5cm 0cm 0cm 3.5cm, clip=true, width=0.1154\linewidth]{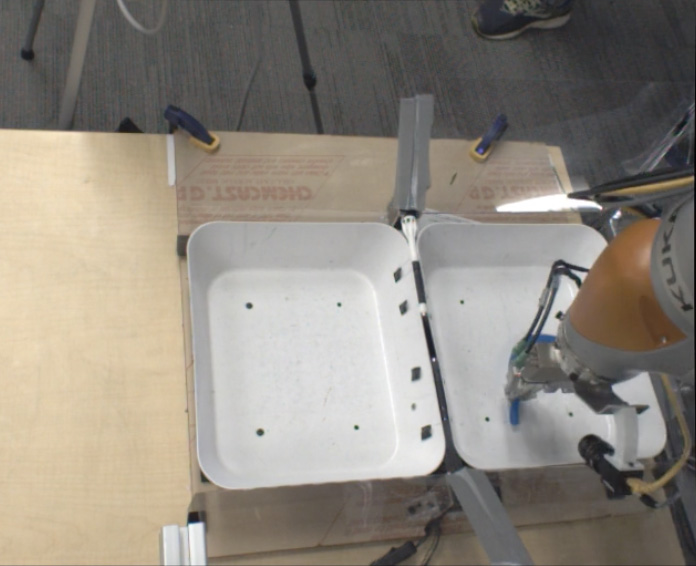} \includegraphics[trim=5cm 0cm 0cm 3.5cm, clip=true,width=0.1154\linewidth]{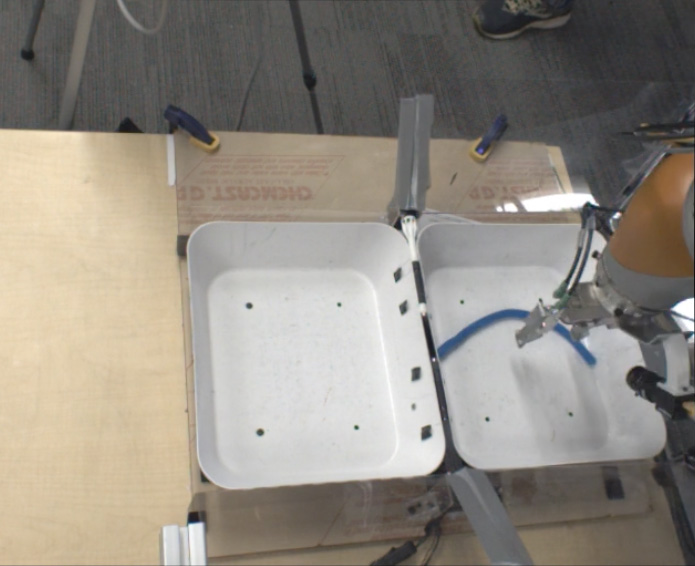}}\\\vspace{-21pt}

\setcounter{subfigure}{0}\subfloat[seen][Instance grasping, Rearrangement, Container placing]{\includegraphics[trim=5cm 0cm 0cm 3.5cm, clip=true, width=0.1154\linewidth]{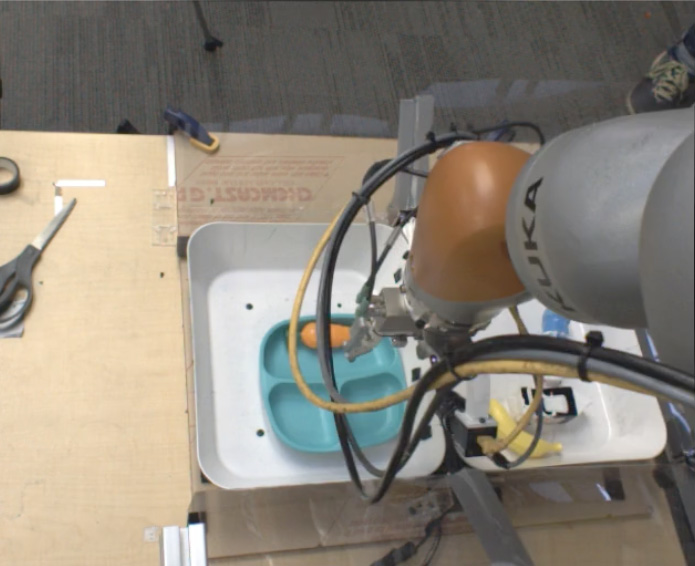}\hspace{2pt}\includegraphics[trim=0cm 3.8cm 0cm -0.1cm, clip=true,width=0.008\linewidth]{images/real_robot/real_camera_sep.jpg}\hspace{2pt}\includegraphics[trim=5cm 0cm 0cm 3.5cm, clip=true, width=0.1154\linewidth]{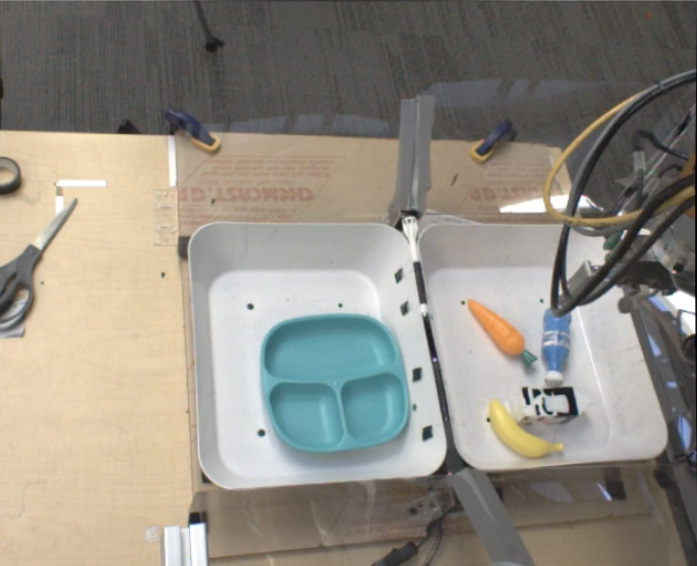} \includegraphics[trim=5cm 0cm 0cm 3.5cm, clip=true, width=0.1154\linewidth]{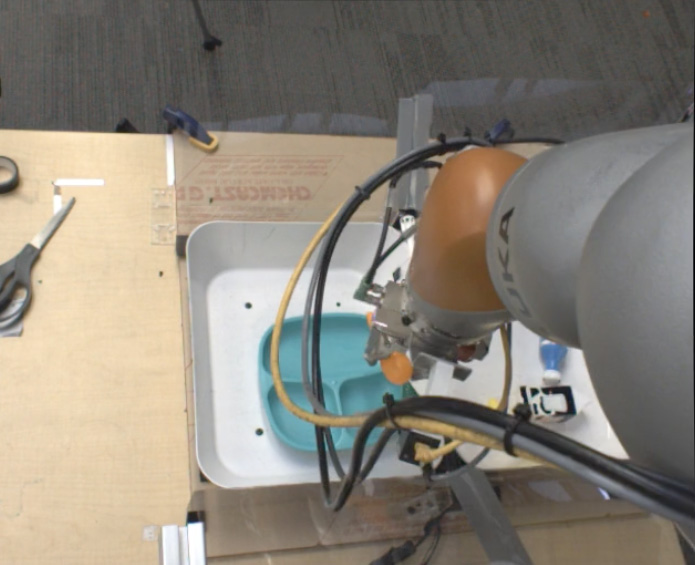} \includegraphics[trim=5cm 0cm 0cm 3.5cm, clip=true,width=0.1154\linewidth]{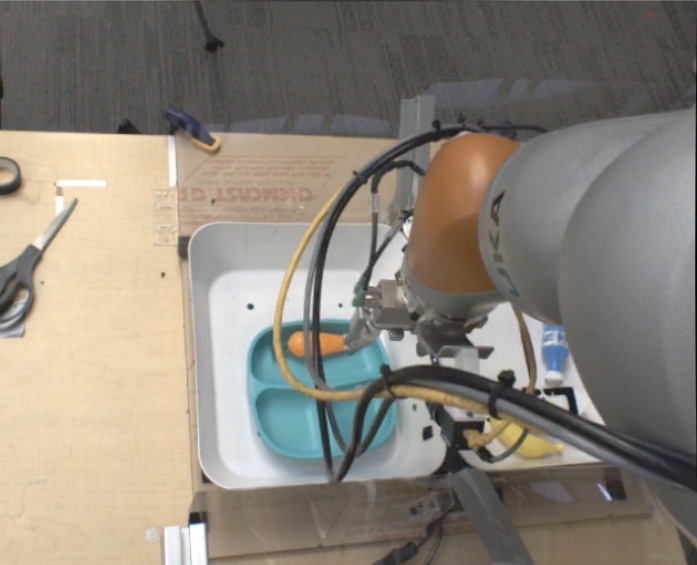}\label{fig:real_robot_train}}\hspace{9pt}
\subfloat[unseen][Unseen object manipulation]{\includegraphics[trim=5cm 0cm 0cm 3.5cm, clip=true, width=0.1154\linewidth]{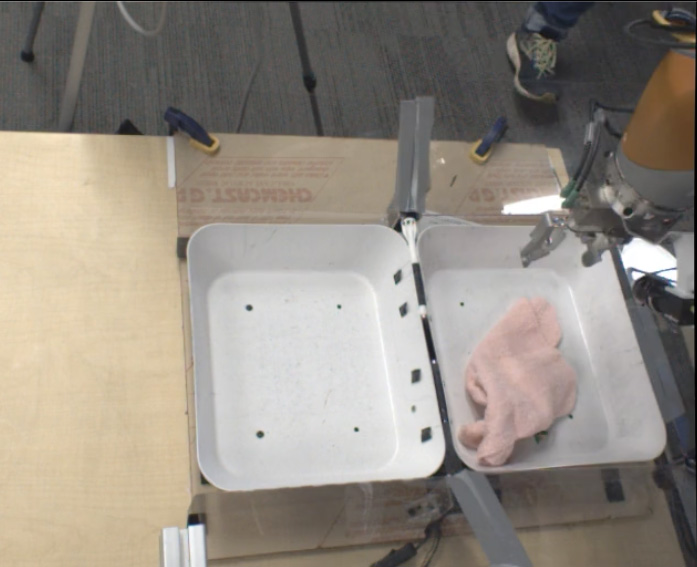}\hspace{2pt}\includegraphics[trim=0cm 3.8cm 0cm -0.1cm, clip=true,width=0.008\linewidth]{images/real_robot/real_camera_sep.jpg}\hspace{2pt}\includegraphics[trim=5cm 0cm 0cm 3.5cm, clip=true, width=0.1154\linewidth]{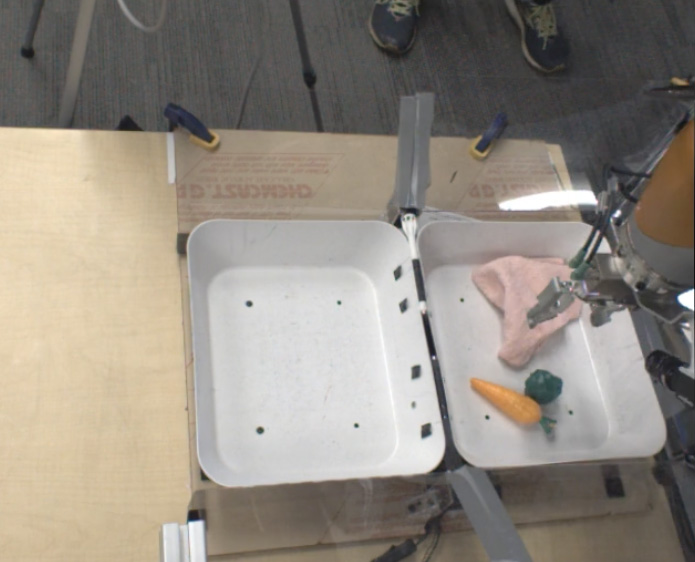} \includegraphics[trim=5cm 0cm 0cm 3.5cm, clip=true, width=0.1154\linewidth]{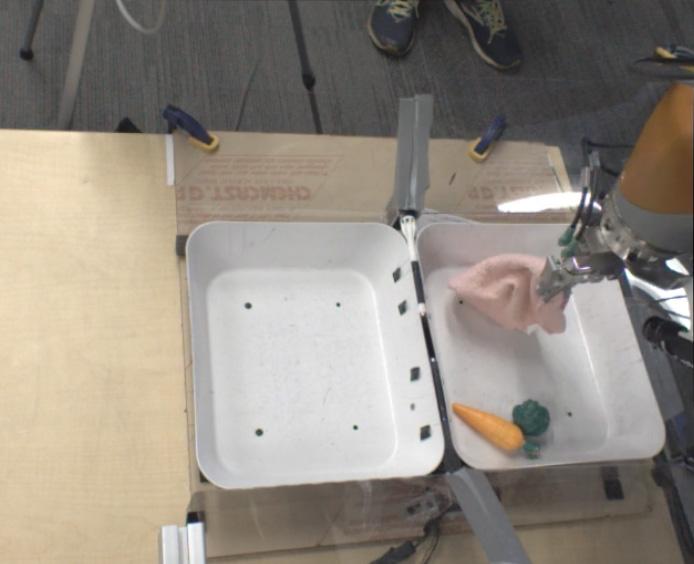} \includegraphics[trim=5cm 0cm 0cm 3.5cm, clip=true,width=0.1154\linewidth]{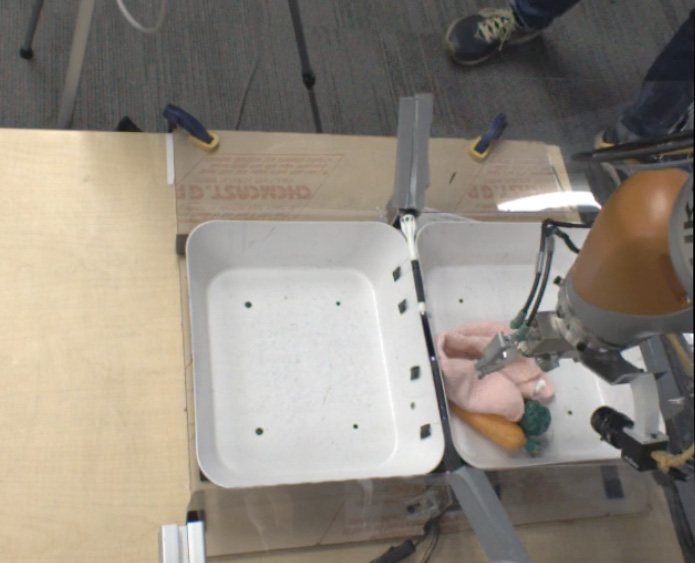}\label{fig:real_robot_unseen}} 
\vspace{-7pt}
\caption{Real robot goal reaching tasks (robot camera view). Robot has to reach the object configuration specified by the goal image.}
\vspace{-5pt}
\label{fig:real_exp_main}
\end{figure*}

\begin{figure*}[t]
\vspace{-5pt}
\captionsetup[subfigure]{justification=centering}
 \hspace*{113pt}
\subfloat{
 \includegraphics[trim=0cm 0cm 0cm 0cm, clip=true, width=0.23\linewidth]{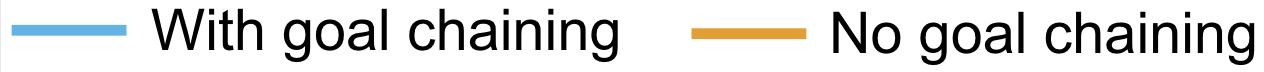}} 
\hspace*{20pt}
\subfloat{
 \includegraphics[trim=0cm 0cm 0cm 0cm, clip=true, width=0.42\linewidth]{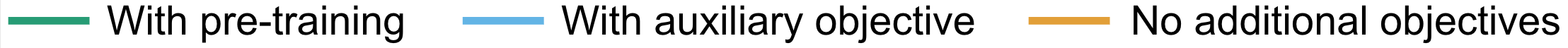}} 
 \\ \vspace{-22pt}
\setcounter{subfigure}{0}
    \centering
\subfloat[real_world_success][Real world goal reaching]{
  \footnotesize{\def\arraystretch{1.3}\begin{tabular}{lc}
  & \vspace{-7pt}\\
\multicolumn{2}{l}{\textbf{Task} \quad \quad \quad \quad\textbf{Success rate}} \\ \hline
Instance grasping & 92\%\\ 
Rearrangement & 74\%\\ 
Container placing & \,66\% \vspace{15pt}\\ 
\end{tabular}}\label{tab:real_results}}
\subfloat[task1][Goal chaining ablation]{
\includegraphics[trim=0cm 0cm 0cm 0cm, clip=true, width=0.24\linewidth,valign=c]{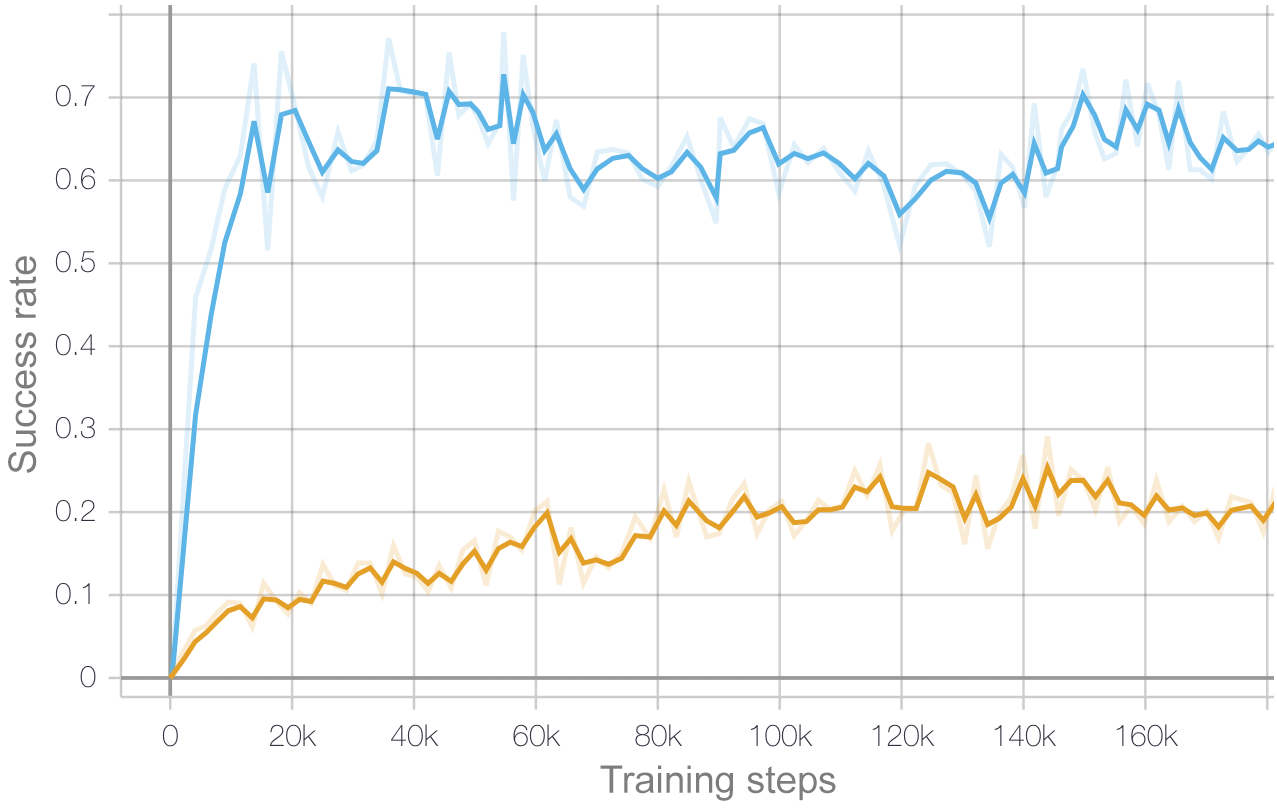}
\label{fig:goal_chaining_comp}}
\subfloat[task11][Bottle grasping]{
\includegraphics[trim=0cm 0cm 0cm 0cm, clip=true, width=0.24\linewidth,valign=c]{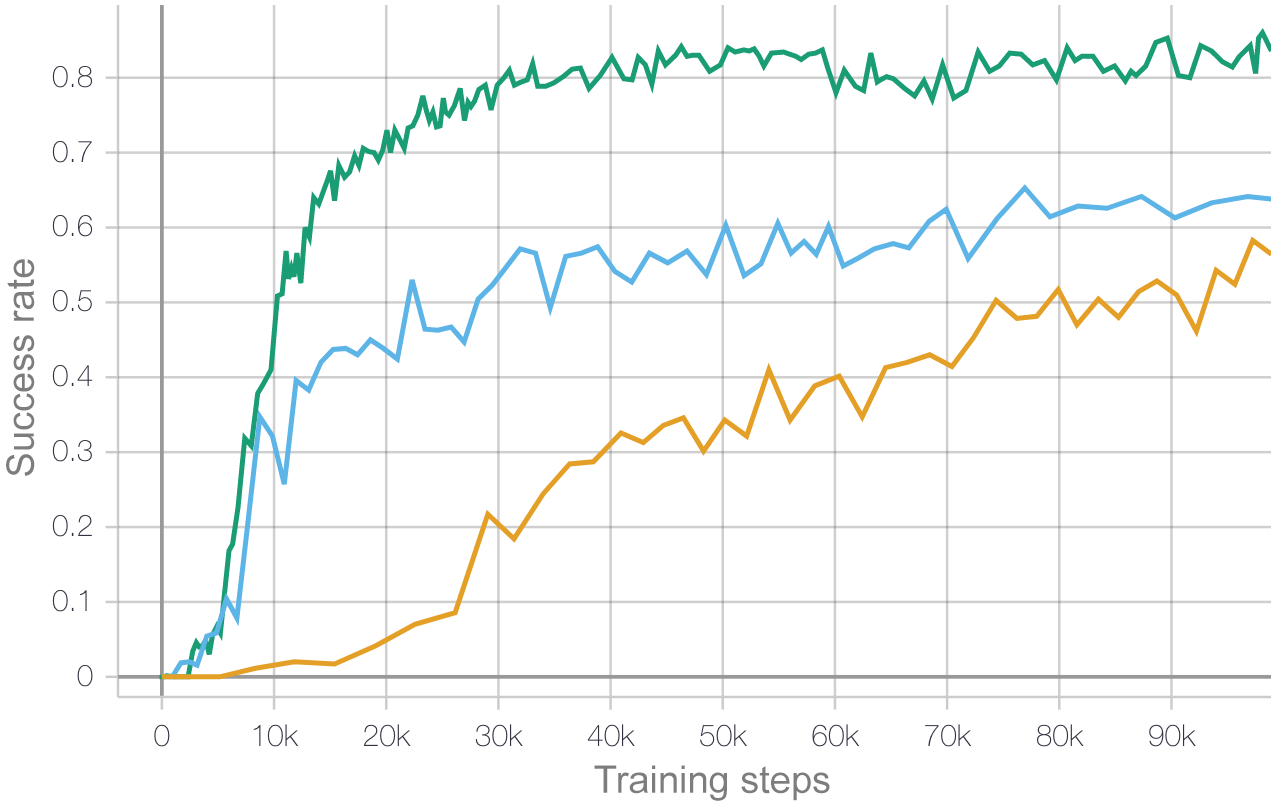}
\label{fig:repr_bottle}
}
\subfloat[task78][Banana grasping]{
\includegraphics[trim=0cm 0cm 0cm 0cm, clip=true, width=0.24\linewidth,valign=c]{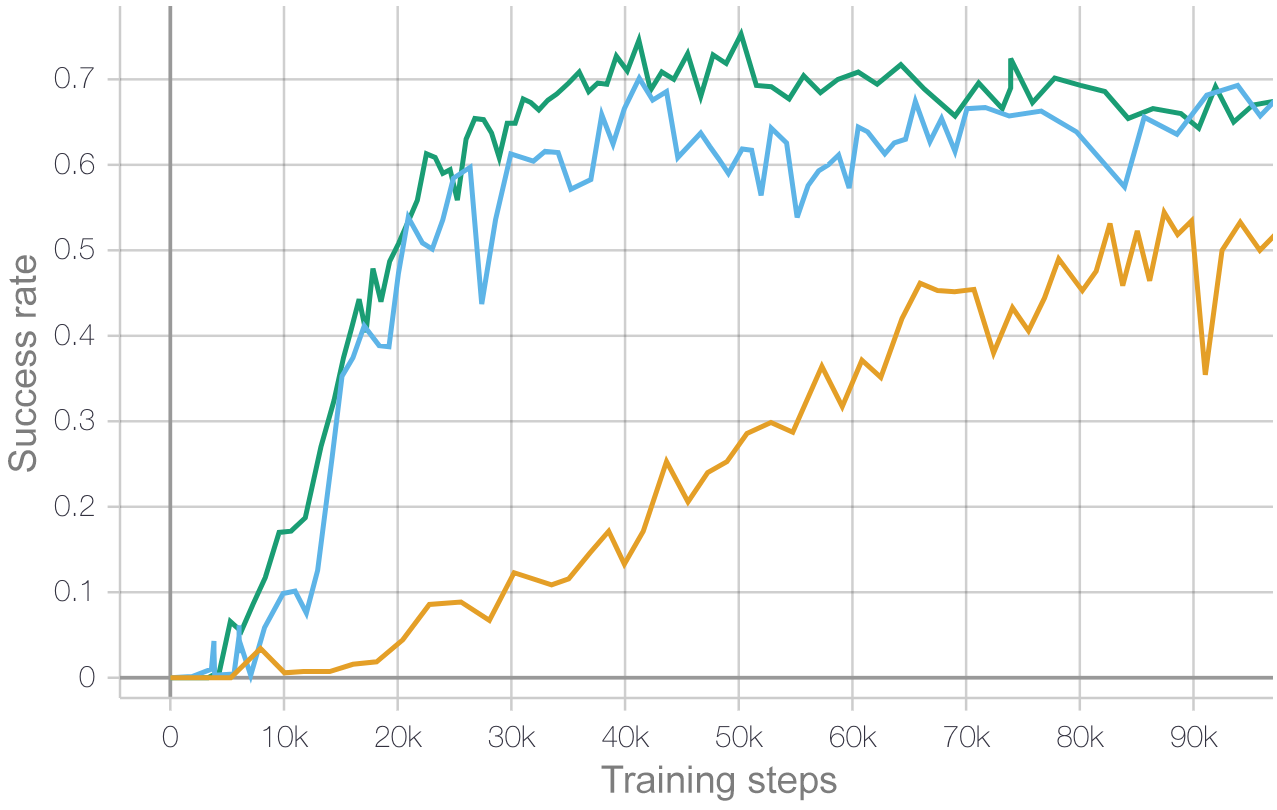}
\label{fig:repr_banana}
}
\vspace{-6pt}
\caption{\textit{a):} Success rates of real world goal reaching skills. \textit{b):} Comparison of training the fixture placing task with separate episodes for grasping and placing, with (blue) and without (yellow) goal chaining enabled. \textit{c) and d)}: Comparison of training the instance grasping tasks using standard QT-Opt without any additional objectives (yellow), pre-training with a goal-conditioned model (green) and 10\% auxiliary objective mix-in (blue).}
\vspace{-5pt}
\end{figure*}

\textbf{Goal-conditioned behavioral cloning (GCBC)}, used in several previous works to learn goal-conditioned policies~\citep{oh2018self, ding2019goal, LynchKXKTLS19,ghosh2019learning}. The policy network follows the same architecture  as in Fig.~\ref{fig:nn}, except with actions as the output.

\textbf{Q-learning with hindsight relabeling}, analogously to hindsight experience replay (HER)~\citep{her} and other prior work~\citep{lin2019reinforcement, qtopt_goal}. Here, we relabel each sub-sequence with the success-reward $R(\tau_{0:i}, s_i)=1$ without any additional regularization. To ensure a fair comparison, the same QT-Opt RL framework is used for training the Q-function.

\textbf{Q-learning with hindsight relabeling and negative random goals}, where in addition to positive relabeling with  $R(\tau_{0:i}, s_i)=1$ for each sub-sequence, we introduce a reward 0 for relabeling with random goals from the data set, such that $R(\tau_{0:i}, g_{rand}) = 0$ with $g_{rand} \sim \mathcal{D}$, which should approximate minimization on $Q(s,a, \tilde{g})$. This corresponds to the standard hindsight experience replay~\cite{her} with additional random goal negatives, which was also used as a baseline by~\citet{EysenbachGLS20}. Here we also use the same QT-Opt RL framework for training the Q-function to ensure fair comparison.

A trained model from each method is evaluated on a variety of goal images from each of the tasks in Fig.~\ref{fig:sim_tasks} over the course of training, with success rates shown in Fig.~\ref{fig:goal_conditioned_comp}. Note that these online evaluations are only used for reporting the results -- the trials are not available for use in training. We also observe that training a unified goal-conditioned policy is harder than training single expert policies with task-specific rewards. Although a goal-conditioned policy covers a larger space of skills, it often leads to a lower performance when evaluated on individual tasks.
The results in~Fig.~\ref{fig:goal_conditioned_comp} show that Actionable Models outperform GCBC on all tasks. Q-learning without any regularization fails to learn any of the tasks, indicating that the Q-function collapses without a presence of negative examples. Q-learning with random goal negatives is only able to learn the simplest of the tasks (block pick-and-place), but still underperforms compared to both Actionable Models and GCBC. With large datasets, random goal negatives can be completely unrelated to the current trajectory, and might even come from a different scene. Such random relabeled goals therefore rarely provide a strong contrastive training signal, making it difficult for the Q-function to learn to distinguish effective actions from ineffective ones.
In prior work that employs online exploration~\citep{her, lin2019reinforcement}, this challenge is mitigated by the fact that the online trials themselves serve as ``hard negatives,'' since they represent failed episodes where the policy was specifically aiming to reach a particular goal. However, this is not available in the offline setting.
Since Actionable Models explicitly minimize the value of unseen actions that are very close to the decision boundary, as noted in Section~\ref{sec:goal_offline}, they do not require careful strategies for sampling ``negative'' goals. Without this regularization, we see that other Q-learning methods are not able to learn any of the skills effectively. 

\vspace{1pt}
\subsection{Real-world visual goal reaching experiments}
\vspace{1pt}
\label{sec:real_robot_exp}

We train a single goal-image conditioned policy on all of the real-world data using the method described in Section~\ref{sec:goal_offline}, combined with goal chaining from Section~\ref{sec:goal_chaining}. Fig.~\ref{fig:real_exp_main} shows examples of skills learned with our method. The qualitative results are best viewed in our supplementary video\footnote{\scriptsize{\mbox{\url{https://actionable-models.github.io}}}}. In order to perform quantitative evaluation and group various goal reaching tasks, we define the following skills:

\textbf{Instance grasping}. Presented with a goal image of an object in the hand, the robot has to grasp the corresponding object from the table. There are at most 8 objects randomly located on the table including the object that has to be grasped.

\textbf{Rearrangement}. Presented with a goal image of a particular object configuration, the robot has to move displaced objects to their goal positions. We mark episodes as successful if the robot moves the objects within 5cm of their goal positions. There are at most 3 objects located on the table and the robot has to move at most 2 objects.

\textbf{Container placing}. Presented with a goal image of an object placed in a container, such as a plate, the robot has to pick up the object and place it into the container. There are at most 4 objected randomly located on the table and one object has to be moved into the container.

To generate goal images, we first place objects in their goal configuration and record the corresponding robot camera image. After that, we reset the objects by moving them to their random initial positions. This setup ensures that the goal image is recorded in the same scene observed during the skill execution. We also include the arm in the goal image as it is usually present in the goal images when performing hindsight relabeling during training.  
Table~\ref{tab:real_results} shows success rates for each of the skills when evaluated on 50 trials for each skill with the best results on instance grasping and the most challenging task being container placing. In addition, we experiment with our model manipulating unseen objects. Fig.~\ref{fig:real_robot_unseen} shows examples of our learned policy manipulating objects that were not present in the training data, such as silverware, a towel, and a rubber cord. This demonstrates that our policy is able to learn general object manipulation skills that can be transferred to novel objects. 

We compare our method to standard Q-learning with hindsight relabeling and no regularization as described in Section~\ref{sec:sim_goal_cond}.
This baseline is not able to learn any of the skills and achieves 0\% success rate as the Q-function collapses without seeing any negative action examples during training.
The breadth of this evaluation provides coverage of a wide range of real world scenes, but unfortunately makes it difficult to compare to the full range of prior methods. We therefore defer a full comparison to prior works to the simulated experiments in Section~\ref{sec:sim_goal_cond}.

\vspace{-2pt}
\subsection{Goal chaining}
\vspace{-1pt}
\label{sec:goal_chaining_exp}
To evaluate the goal chaining method described in Section~\ref{sec:goal_chaining}, we first ablate the goal chaining component in simulation. We split the fixture placing task trajectories (Fig.~\ref{fig:sim_tasks}) in our dataset  into two separate trajectories: the grasping part and the placing part. We shuffle the dataset, meaning that these trajectories are not connected in any way. We then train the goal-conditioned Q-function with and without goal chaining, and evaluate by conditioning on the final goal of the fixture placing task, where the object is fully placed in the fixture. The results are shown in Fig.~\ref{fig:goal_chaining_comp}. When we disable goal chaining, performance drops significantly, as the Q-function is only trained on goals within single episodes, which does not include performing the fixture placing task all the way from the beginning of grasping to the end. With goal chaining, the model can learn to reach the placing goal across two episodes and successfully perform the task.

The above setting is in some sense ``ideal'' for goal chaining. To evaluate goal chaining under more realistic conditions, we also perform an ablation study with real robots by training our goal-conditioned model on the data from Section~\ref{sec:real_robot_exp}, but with goal chaining disabled. We observe a performance drop from 66\% down to 8\% success rate on the longer-horizon container placing tasks, which are more likely to be spread out across multiple episodes. This indicates that by enabling goal chaining we can better utilize the dataset, especially for longer-horizon tasks that might require combining several episodes.

\vspace{-1pt}
\subsection{Representation learning}
\vspace{-1pt}
\label{sec:exp_repr_learning}
\begin{table}[t]
\footnotesize
\begin{center}
\def\arraystretch{1.2}
\setlength{\tabcolsep}{5pt}
\begin{tabular}{lcc}
  & \vspace{-8pt}\\
\textbf{Task} & \textbf{No pre-training}  & \textbf{With pre-training} \\ \hline
Grasp box & 0\%&27\%\\ 
Grasp banana & 4\% & 20\%\\ 
Grasp milk & 1\% & 20\%\\ 
\end{tabular}
\end{center}
\vspace{-10pt}
\caption{Success rates of learning real world instance grasping tasks from a small amount of data with task-specific rewards: without any pre-training, pre-training with a goal-conditioned model and fine-tuning with task-specific rewards.}
\label{tab:real_fine_tune}
\vspace{-10pt}
\end{table}

Lastly, we analyze how Actionable Models can help with pre-training rich representations or provide an auxiliary objective for standard RL, as described in Sections~\ref{sec:pretrain} and~\ref{sec:aux}.
First, we perform simulated experiments on two instance grasping versions of the food object grasping task (Fig.~\ref{fig:sim_tasks} on the right), which are harder for the standard QT-Opt policy to learn. The goal is to grasp a specific object, such as a bottle or a banana, which requires identifying the object and then singulating it from clutter before grasping and lifting. For pre-training, we use the goal-conditioned model trained on 4 tasks, as in Section~\ref{sec:sim_goal_cond}, which we then fine-tune with additional online data collection and a standard task reward. During fine-tuning, the goal-image input is set to all zeros.
To add the auxiliary goal reaching objective, we mix in relabeled episodes with a probability of 10\% during the online training of the task as described in Section~\ref{sec:aux}. Figures~\ref{fig:repr_bottle} and~\ref{fig:repr_banana} compare learning with pre-training and an auxiliary objective to standard online RL with QT-Opt without any additional objectives.
The results show that both pre-training and the auxiliary objective significantly speed up the training of the downstream tasks, suggesting that functional knowledge acquired by Actionable Models is beneficial for learning other robotic skills.

In our real world experiments, we first pre-train a goal-conditioned model on the data from Section~\ref{sec:real_robot_exp} and then fine-tune it on a small amount of data ($<$10K episodes) for three instance grasping tasks with task-specific rewards, as shown in Table~\ref{tab:real_fine_tune}. Without any pre-training, the standard offline QT-Opt is not able to learn these tasks and achieves less than 5\% success, while fine-tuning a pre-trained model enables reaching at least 20\% success rate on all tasks. This confirms that representations learned by pre-training with a goal-reaching objective also improve learning performance in the real world.
\vspace*{2pt}
\section{Discussion and Future Work}
\vspace{1pt}
\label{sec:conclusion}

In this work, we showed that it is possible to learn diverse  skills on real robots with high-dimensional goal images in a completely offline setting.
We developed a method that enables doing this with goal-conditioned Q-functions, which are particularly challenging to train in an offline setting. We presented a regularization method that ensures that the Q-values of unseen actions are not overestimated, and a goal chaining technique that enables reaching goals across multiple episodes. We also showed that downstream RL tasks can be learned more efficiently by either fine-tuning pre-trained goal-conditioned models, or through a goal-reaching auxiliary objective during training.

While our approach enables learning diverse goal-reaching tasks and accelerates acquisition of downstream behaviors, it does have a number of limitations.
First, specifying a task to the goal-conditioned Q-function requires providing a suitable goal image at test-time, which should be consistent with the current scene. This limits the ability to specify general tasks, e.g. using goal images from a different scene or commanding the robot to grasp a particular type of object instead of reaching a goal image. Joint training and pre-training with fine-tuning can sidestep this limitation by using additional general task rewards, as discussed in Section~\ref{sec:exp_repr_learning}.
An exciting direction for future work is to  study how representation learning and goal embeddings can further alleviate this limitation for general goal-conditioned policies.
Second, reaching some goals requires reasoning over extended horizons (e.g., moving multiple objects to desired locations). 
Currently, our approach  cannot reposition a large number of objects in a single episode. Combining planning algorithms with goal-conditioned RL, as in some recent works~\cite{eysenbach2019search,ichter2020broadlyexploring}, could be an exciting direction for addressing this limitation.

Going forward, we believe that our work suggests a practical and broadly applicable technique for generalizable robotic learning: by training goal-conditioned Q-functions from diverse multi-task offline data, robots can acquire a broad knowledge about the world. This knowledge, in turn, can then be used to directly perform goal-reaching tasks or as a way to accelerate training downstream policies with standard online reinforcement learning methods. 

\vspace{1pt}
\section*{Acknowledgements}
\vspace{1pt}
We would like to thank Benjamin Swanson, Josh Weaver, Noah Brown, Khem Holden, Linda  Luu  and  Brandon  Kinman for  their robot  operation  support, Tom Small for help with videos and project media, Julian  Ibarz, Kanishka  Rao, and  Vincent  Vanhoucke for their managerial support, and all of  the  Robotics  at Google team for their support throughout this project.

\bibliographystyle{icml2021} 
\bibliography{references}

\begin{thebibliography}{50}
\providecommand{\natexlab}[1]{#1}
\providecommand{\url}[1]{\texttt{#1}}
\expandafter\ifx\csname urlstyle\endcsname\relax
  \providecommand{\doi}[1]{doi: #1}\else
  \providecommand{\doi}{doi: \begingroup \urlstyle{rm}\Url}\fi

\bibitem[Andrychowicz et~al.(2017)Andrychowicz, Crow, Ray, Schneider, Fong,
  Welinder, McGrew, Tobin, Abbeel, and Zaremba]{her}
Andrychowicz, M., Crow, D., Ray, A., Schneider, J., Fong, R., Welinder, P.,
  McGrew, B., Tobin, J., Abbeel, P., and Zaremba, W.
\newblock Hindsight experience replay.
\newblock In \emph{NeurIPS}, pp.\  5048--5058, 2017.

\bibitem[Barreto et~al.(2016)Barreto, Dabney, Munos, Hunt, Schaul, Van~Hasselt,
  and Silver]{barreto2016successor}
Barreto, A., Dabney, W., Munos, R., Hunt, J.~J., Schaul, T., Van~Hasselt, H.,
  and Silver, D.
\newblock Successor features for transfer in reinforcement learning.
\newblock \emph{arXiv}, abs/1606.05312, 2016.

\bibitem[Bauza \& Rodriguez(2017)Bauza and Rodriguez]{bauza2017probabilistic}
Bauza, M. and Rodriguez, A.
\newblock A probabilistic data-driven model for planar pushing.
\newblock In \emph{2017 IEEE International Conference on Robotics and
  Automation (ICRA)}, pp.\  3008--3015. IEEE, 2017.

\bibitem[Brown et~al.(2020)Brown, Mann, Ryder, Subbiah, Kaplan, Dhariwal,
  Neelakantan, Shyam, Sastry, Askell, Agarwal, Herbert{-}Voss, Krueger,
  Henighan, Child, Ramesh, Ziegler, Wu, Winter, Hesse, Chen, Sigler, Litwin,
  Gray, Chess, Clark, Berner, McCandlish, Radford, Sutskever, and Amodei]{gpt3}
Brown, T.~B., Mann, B., Ryder, N., Subbiah, M., Kaplan, J., Dhariwal, P.,
  Neelakantan, A., Shyam, P., Sastry, G., Askell, A., Agarwal, S.,
  Herbert{-}Voss, A., Krueger, G., Henighan, T., Child, R., Ramesh, A.,
  Ziegler, D.~M., Wu, J., Winter, C., Hesse, C., Chen, M., Sigler, E., Litwin,
  M., Gray, S., Chess, B., Clark, J., Berner, C., McCandlish, S., Radford, A.,
  Sutskever, I., and Amodei, D.
\newblock Language models are few-shot learners.
\newblock \emph{arXiv}, abs/2005.14165, 2020.

\bibitem[Buckman et~al.(2018)Buckman, Hafner, Tucker, Brevdo, and
  Lee]{buckman2018sample}
Buckman, J., Hafner, D., Tucker, G., Brevdo, E., and Lee, H.
\newblock Sample-efficient reinforcement learning with stochastic ensemble
  value expansion.
\newblock \emph{arXiv}, abs/1807.01675, 2018.

\bibitem[Chua et~al.(2018)Chua, Calandra, McAllister, and Levine]{chua2018deep}
Chua, K., Calandra, R., McAllister, R., and Levine, S.
\newblock Deep reinforcement learning in a handful of trials using
  probabilistic dynamics models.
\newblock \emph{arXiv}, abs/1805.12114, 2018.

\bibitem[Dayan(1993)]{dayan1993improving}
Dayan, P.
\newblock Improving generalization for temporal difference learning: The
  successor representation.
\newblock \emph{Neural Computation}, 5\penalty0 (4):\penalty0 613--624, 1993.

\bibitem[Deisenroth \& Rasmussen(2011)Deisenroth and
  Rasmussen]{deisenroth2011pilco}
Deisenroth, M. and Rasmussen, C.~E.
\newblock Pilco: A model-based and data-efficient approach to policy search.
\newblock In \emph{International Conference on Machine Learning}, pp.\
  465--472, 2011.

\bibitem[Denton \& Fergus(2018)Denton and Fergus]{denton2018stochastic}
Denton, E. and Fergus, R.
\newblock Stochastic video generation with a learned prior.
\newblock In \emph{International Conference on Machine Learning}, pp.\
  1174--1183. PMLR, 2018.

\bibitem[Devlin et~al.(2018)Devlin, Chang, Lee, and Toutanova]{bert}
Devlin, J., Chang, M., Lee, K., and Toutanova, K.
\newblock {BERT:} pre-training of deep bidirectional transformers for language
  understanding.
\newblock \emph{arXiv}, abs/1810.04805, 2018.

\bibitem[Ding et~al.(2019)Ding, Florensa, Phielipp, and Abbeel]{ding2019goal}
Ding, Y., Florensa, C., Phielipp, M., and Abbeel, P.
\newblock Goal-conditioned imitation learning.
\newblock \emph{arXiv}, abs/1906.05838, 2019.

\bibitem[Eysenbach et~al.(2019)Eysenbach, Salakhutdinov, and
  Levine]{eysenbach2019search}
Eysenbach, B., Salakhutdinov, R., and Levine, S.
\newblock Search on the replay buffer: Bridging planning and reinforcement
  learning.
\newblock \emph{arXiv}, abs/1906.05253, 2019.

\bibitem[Eysenbach et~al.(2020)Eysenbach, Geng, Levine, and
  Salakhutdinov]{EysenbachGLS20}
Eysenbach, B., Geng, X., Levine, S., and Salakhutdinov, R.~R.
\newblock Rewriting history with inverse {RL:} hindsight inference for policy
  improvement.
\newblock In \emph{NeurIPS}, 2020.

\bibitem[Finn \& Levine(2017)Finn and Levine]{finn2017deep}
Finn, C. and Levine, S.
\newblock Deep visual foresight for planning robot motion.
\newblock In \emph{2017 IEEE International Conference on Robotics and
  Automation (ICRA)}, pp.\  2786--2793. IEEE, 2017.

\bibitem[Fujita et~al.(2020)Fujita, Uenishi, Ummadisingu, Nagarajan, Masuda,
  and Castro]{qtopt_goal}
Fujita, Y., Uenishi, K., Ummadisingu, A., Nagarajan, P., Masuda, S., and
  Castro, M.~Y.
\newblock Distributed reinforcement learning of targeted grasping with active
  vision for mobile manipulators.
\newblock \emph{arXiv}, abs/2007.08082, 2020.

\bibitem[Ghosh et~al.(2019)Ghosh, Gupta, Fu, Reddy, Devin, Eysenbach, and
  Levine]{ghosh2019learning}
Ghosh, D., Gupta, A., Fu, J., Reddy, A., Devin, C., Eysenbach, B., and Levine,
  S.
\newblock Learning to reach goals without reinforcement learning.
\newblock \emph{arXiv}, abs/1912.06088, 2019.

\bibitem[Greydanus \& Olah(2019)Greydanus and Olah]{greydanus2019the}
Greydanus, S. and Olah, C.
\newblock The paths perspective on value learning.
\newblock \emph{Distill}, 2019.
\newblock \doi{10.23915/distill.00020}.
\newblock https://distill.pub/2019/paths-perspective-on-value-learning.

\bibitem[Hafner et~al.(2019)Hafner, Lillicrap, Fischer, Villegas, Ha, Lee, and
  Davidson]{hafner2019learning}
Hafner, D., Lillicrap, T., Fischer, I., Villegas, R., Ha, D., Lee, H., and
  Davidson, J.
\newblock Learning latent dynamics for planning from pixels.
\newblock In \emph{International Conference on Machine Learning}, pp.\
  2555--2565. PMLR, 2019.

\bibitem[Ichter et~al.(2020)Ichter, Sermanet, and
  Lynch]{ichter2020broadlyexploring}
Ichter, B., Sermanet, P., and Lynch, C.
\newblock Broadly-exploring, local-policy trees for long-horizon task planning.
\newblock \emph{arXiv}, abs/2010.06491, 2020.

\bibitem[Janz et~al.(2018)Janz, Hron, Mazur, Hofmann, Hern{\'a}ndez-Lobato, and
  Tschiatschek]{janz2018successor}
Janz, D., Hron, J., Mazur, P., Hofmann, K., Hern{\'a}ndez-Lobato, J.~M., and
  Tschiatschek, S.
\newblock Successor uncertainties: exploration and uncertainty in temporal
  difference learning.
\newblock \emph{arXiv}, abs/1810.06530, 2018.

\bibitem[Kadian et~al.(2020)Kadian, Truong, Gokaslan, Clegg, Wijmans, Lee,
  Savva, Chernova, and Batra]{kadian2020sim2real}
Kadian, A., Truong, J., Gokaslan, A., Clegg, A., Wijmans, E., Lee, S., Savva,
  M., Chernova, S., and Batra, D.
\newblock Sim2real predictivity: Does evaluation in simulation predict
  real-world performance?
\newblock \emph{IEEE Robotics and Automation Letters}, 5\penalty0 (4):\penalty0
  6670--6677, 2020.

\bibitem[Kaelbling(1993)]{Kaelbling93}
Kaelbling, L.~P.
\newblock Learning to achieve goals.
\newblock In Bajcsy, R. (ed.), \emph{IJCAI}, pp.\  1094--1099. Morgan Kaufmann,
  1993.
\newblock ISBN 1-55860-300-X.

\bibitem[Kaiser et~al.(2019)Kaiser, Babaeizadeh, Milos, Osinski, Campbell,
  Czechowski, Erhan, Finn, Kozakowski, Levine, et~al.]{kaiser2019model}
Kaiser, L., Babaeizadeh, M., Milos, P., Osinski, B., Campbell, R.~H.,
  Czechowski, K., Erhan, D., Finn, C., Kozakowski, P., Levine, S., et~al.
\newblock Model-based reinforcement learning for atari.
\newblock \emph{arXiv}, abs/1903.00374, 2019.

\bibitem[Kalashnikov et~al.(2018)Kalashnikov, Irpan, Pastor, Ibarz, Herzog,
  Jang, Quillen, Holly, Kalakrishnan, Vanhoucke, and Levine]{qtopt}
Kalashnikov, D., Irpan, A., Pastor, P., Ibarz, J., Herzog, A., Jang, E.,
  Quillen, D., Holly, E., Kalakrishnan, M., Vanhoucke, V., and Levine, S.
\newblock Qt-opt: Scalable deep reinforcement learning for vision-based robotic
  manipulation.
\newblock \emph{arXiv}, abs/1806.10293, 2018.

\bibitem[Kalashnikov et~al.(2021)Kalashnikov, Varley, Chebotar, Swanson,
  Jonschkowski, Finn, Levine, and Hausman]{mtopt2021arxiv}
Kalashnikov, D., Varley, J., Chebotar, Y., Swanson, B., Jonschkowski, R., Finn,
  C., Levine, S., and Hausman, K.
\newblock Mt-opt: Continuous multi-task robotic reinforcement learning at
  scale.
\newblock \emph{arXiv}, abs/2104.08212, 2021.

\bibitem[Kumar et~al.(2019{\natexlab{a}})Kumar, Fu, Soh, Tucker, and
  Levine]{kumar2019stabilizing}
Kumar, A., Fu, J., Soh, M., Tucker, G., and Levine, S.
\newblock Stabilizing off-policy q-learning via bootstrapping error reduction.
\newblock In \emph{NeurIPS}, pp.\  11761--11771, 2019{\natexlab{a}}.

\bibitem[Kumar et~al.(2020)Kumar, Zhou, Tucker, and Levine]{cql}
Kumar, A., Zhou, A., Tucker, G., and Levine, S.
\newblock Conservative q-learning for offline reinforcement learning.
\newblock \emph{arXiv}, abs/2006.04779, 2020.

\bibitem[Kumar et~al.(2019{\natexlab{b}})Kumar, Babaeizadeh, Erhan, Finn,
  Levine, Dinh, and Kingma]{kumar2019videoflow}
Kumar, M., Babaeizadeh, M., Erhan, D., Finn, C., Levine, S., Dinh, L., and
  Kingma, D.
\newblock Videoflow: A flow-based generative model for video.
\newblock \emph{arXiv}, abs/1903.01434, 2019{\natexlab{b}}.

\bibitem[Lange et~al.(2012)Lange, Gabel, and Riedmiller]{lange2012batch}
Lange, S., Gabel, T., and Riedmiller, M.
\newblock Batch reinforcement learning.
\newblock In \emph{Reinforcement learning}, pp.\  45--73. Springer, 2012.

\bibitem[Laroche et~al.(2017)Laroche, Trichelair, and Combes]{laroche2017safe}
Laroche, R., Trichelair, P., and Combes, R. T.~d.
\newblock Safe policy improvement with baseline bootstrapping.
\newblock \emph{arXiv}, abs/1712.06924, 2017.

\bibitem[Lin et~al.(2019)Lin, Baweja, and Held]{lin2019reinforcement}
Lin, X., Baweja, H.~S., and Held, D.
\newblock Reinforcement learning without ground-truth state.
\newblock \emph{arXiv}, abs/1905.07866, 2019.

\bibitem[Lynch et~al.(2019)Lynch, Khansari, Xiao, Kumar, Tompson, Levine, and
  Sermanet]{LynchKXKTLS19}
Lynch, C., Khansari, M., Xiao, T., Kumar, V., Tompson, J., Levine, S., and
  Sermanet, P.
\newblock Learning latent plans from play.
\newblock In \emph{CoRL}, volume 100 of \emph{Proceedings of Machine Learning
  Research}, pp.\  1113--1132. PMLR, 2019.

\bibitem[Mathieu et~al.(2015)Mathieu, Couprie, and Lecun]{mathieu2015video}
Mathieu, M., Couprie, C., and Lecun, Y.
\newblock Deep multi-scale video prediction beyond mean square error.
\newblock \emph{arXiv}, abs/1511.05440, 2015.

\bibitem[Mnih et~al.(2015)Mnih, Kavukcuoglu, Silver, Rusu, Veness, Bellemare,
  Graves, Riedmiller, Fidjeland, Ostrovski, Petersen, Beattie, Sadik,
  Antonoglou, King, Kumaran, Wierstra, Legg, and Hassabis]{mnih2015humanlevel}
Mnih, V., Kavukcuoglu, K., Silver, D., Rusu, A.~A., Veness, J., Bellemare,
  M.~G., Graves, A., Riedmiller, M., Fidjeland, A.~K., Ostrovski, G., Petersen,
  S., Beattie, C., Sadik, A., Antonoglou, I., King, H., Kumaran, D., Wierstra,
  D., Legg, S., and Hassabis, D.
\newblock Human-level control through deep reinforcement learning.
\newblock \emph{Nature}, 518\penalty0 (7540):\penalty0 529--533, 2015.
\newblock ISSN 14764687.

\bibitem[Nagabandi et~al.(2020)Nagabandi, Konolige, Levine, and
  Kumar]{nagabandi2020deep}
Nagabandi, A., Konolige, K., Levine, S., and Kumar, V.
\newblock Deep dynamics models for learning dexterous manipulation.
\newblock In \emph{Conference on Robot Learning}, pp.\  1101--1112. PMLR, 2020.

\bibitem[Nair et~al.(2018)Nair, Pong, Dalal, Bahl, Lin, and
  Levine]{NairPDBLL18}
Nair, A., Pong, V., Dalal, M., Bahl, S., Lin, S., and Levine, S.
\newblock Visual reinforcement learning with imagined goals.
\newblock In \emph{NeurIPS}, pp.\  9209--9220, 2018.

\bibitem[Oh et~al.(2018)Oh, Guo, Singh, and Lee]{oh2018self}
Oh, J., Guo, Y., Singh, S., and Lee, H.
\newblock Self-imitation learning.
\newblock In \emph{International Conference on Machine Learning}, pp.\
  3878--3887. PMLR, 2018.

\bibitem[Pong et~al.(2018)Pong, Gu, Dalal, and Levine]{pong2018temporal}
Pong, V., Gu, S., Dalal, M., and Levine, S.
\newblock Temporal difference models: Model-free deep rl for model-based
  control.
\newblock \emph{arXiv}, abs/1802.09081, 2018.

\bibitem[Rubinstein \& Kroese(2004)Rubinstein and Kroese]{cem}
Rubinstein, R.~Y. and Kroese, D.~P.
\newblock \emph{The Cross Entropy Method: A Unified Approach To Combinatorial
  Optimization, Monte-Carlo Simulation (Information Science and Statistics)}.
\newblock Springer-Verlag, Berlin, Heidelberg, 2004.
\newblock ISBN 038721240X.

\bibitem[Schaul et~al.(2015)Schaul, Horgan, Gregor, and Silver]{uvf}
Schaul, T., Horgan, D., Gregor, K., and Silver, D.
\newblock Universal value function approximators.
\newblock In Bach, F. and Blei, D. (eds.), \emph{International Conference on
  Machine Learning}, volume~37 of \emph{Proceedings of Machine Learning
  Research}, pp.\  1312--1320, Lille, France, 07--09 Jul 2015. PMLR.

\bibitem[Schroecker \& Isbell(2020)Schroecker and
  Isbell]{schroecker2020universal}
Schroecker, Y. and Isbell, C.
\newblock Universal value density estimation for imitation learning and
  goal-conditioned reinforcement learning.
\newblock \emph{arXiv}, abs/2002.06473, 2020.

\bibitem[Singh et~al.(2020)Singh, Yu, Yang, Zhang, Kumar, and
  Levine]{singh2020cog}
Singh, A., Yu, A., Yang, J., Zhang, J., Kumar, A., and Levine, S.
\newblock Cog: Connecting new skills to past experience with offline
  reinforcement learning.
\newblock \emph{arXiv}, abs/2010.14500, 2020.

\bibitem[Sun et~al.(2019)Sun, Li, Liu, Lin, and Zhou]{sun2019policy}
Sun, H., Li, Z., Liu, X., Lin, D., and Zhou, B.
\newblock Policy continuation with hindsight inverse dynamics.
\newblock \emph{arXiv}, abs/1910.14055, 2019.

\bibitem[Sutton(1988)]{sutton1988learning}
Sutton, R.~S.
\newblock Learning to predict by the methods of temporal differences.
\newblock \emph{Machine learning}, 3\penalty0 (1):\penalty0 9--44, 1988.

\bibitem[Sutton et~al.(2011)Sutton, Modayil, Delp, Degris, Pilarski, White, and
  Precup]{sutton2011horde}
Sutton, R.~S., Modayil, J., Delp, M., Degris, T., Pilarski, P.~M., White, A.,
  and Precup, D.
\newblock Horde: A scalable real-time architecture for learning knowledge from
  unsupervised sensorimotor interaction.
\newblock In \emph{The 10th International Conference on Autonomous Agents and
  Multiagent Systems-Volume 2}, pp.\  761--768, 2011.

\bibitem[Wang \& Ba(2019)Wang and Ba]{wang2019exploring}
Wang, T. and Ba, J.
\newblock Exploring model-based planning with policy networks.
\newblock \emph{arXiv}, abs/1906.08649, 2019.

\bibitem[Watter et~al.(2015)Watter, Springenberg, Boedecker, and
  Riedmiller]{watter2015embed}
Watter, M., Springenberg, J.~T., Boedecker, J., and Riedmiller, M.
\newblock Embed to control: A locally linear latent dynamics model for control
  from raw images.
\newblock \emph{arXiv}, abs/1506.07365, 2015.

\bibitem[Williams et~al.(2015)Williams, Aldrich, and
  Theodorou]{williams2015model}
Williams, G., Aldrich, A., and Theodorou, E.
\newblock Model predictive path integral control using covariance variable
  importance sampling.
\newblock \emph{arXiv}, abs/1509.01149, 2015.

\bibitem[Xu \& Denil(2019)Xu and Denil]{xu2019positive}
Xu, D. and Denil, M.
\newblock Positive-unlabeled reward learning.
\newblock \emph{arXiv}, abs/1911.00459, 2019.

\bibitem[Zolna et~al.(2020)Zolna, Novikov, Konyushkova, Gulcehre, Wang, Aytar,
  Denil, de~Freitas, and Reed]{zolna2020offline}
Zolna, K., Novikov, A., Konyushkova, K., Gulcehre, C., Wang, Z., Aytar, Y.,
  Denil, M., de~Freitas, N., and Reed, S.
\newblock Offline learning from demonstrations and unlabeled experience.
\newblock \emph{arXiv}, abs/2011.13885, 2020.

\end{thebibliography}

\end{document}